\documentclass[journal]{IEEEtran}
\usepackage{graphicx}
\input{psfig.sty}
\input{epsf.sty}
\usepackage{amsmath}
\usepackage{amssymb}
\usepackage{amsfonts}
\usepackage{amscd}
\usepackage{mathrsfs}
\usepackage{psfrag}
\usepackage{caption}
\usepackage{subcaption}
\usepackage{breqn}
\usepackage{algorithm,caption}
\usepackage[noend]{algpseudocode}
\usepackage{color}

\newcommand{\mbx}{\mathbf{x}}
\newcommand{\mbA}{\mathbf{A}}
\newcommand{\mbH}{\mathbf{H}}
\newcommand{\mby}{\mathbf{y}}
\newcommand{\mbe}{\mathbf{e}}

\newcommand{\mbz}{\mathbf{z}}

\newcommand{\T}{\mathcal{T}}

\newcommand{\supp}{\ensuremath{\texttt{supp}}}

\newtheorem{theorem}{Theorem}
\newtheorem{lemma}{Lemma}
\newtheorem{proposition}{Proposition}

\newtheorem{definition}{Definition}
\newtheorem{assumption}{Assumption}

\newcommand{\qed}{\nobreak \ifvmode \relax \else
      \ifdim\lastskip<1.5em \hskip-\lastskip
      \hskip1.5em plus0em minus0.5em \fi \nobreak
      \vrule height0.75em width0.5em depth0.25em\fi}

\DeclareMathOperator*{\argmin}{arg\,min}
\newcommand*{\argminl}{\argmin\limits}

\newcommand*{\QEDB}{\hfill\ensuremath{\square}}%
\begin{document}
%
\title{Locally Convex Sparse Learning over Networks}


%
\author{\IEEEauthorblockN{Ahmed Zaki\IEEEauthorrefmark{1}, Saikat Chatterjee\IEEEauthorrefmark{1}, Partha P. Mitra\IEEEauthorrefmark{2} and Lars K. Rasmussen\IEEEauthorrefmark{1}} \\
\IEEEauthorblockA{\IEEEauthorrefmark{1} School of Electrical Engineering, KTH Royal Institute of Technology, Stockholm, Sweden} \\
\IEEEauthorblockA{\IEEEauthorrefmark{2} Cold Spring Harbor Laboratory, 1 Bungtown Road, New York, USA}}


\maketitle

\begin{abstract}
We consider a distributed learning setup where a sparse signal is estimated over a network. Our main interest is to save communication resource for information exchange over the network and reduce processing time. Each node of the network uses a convex optimization based algorithm that provides a locally optimum solution for that node. The nodes exchange their signal estimates over the network in order to refine their local estimates. At a node, the optimization algorithm is based on an $\ell_1$-norm minimization with appropriate modifications to promote sparsity as well as to include influence of estimates from neighboring nodes. Our expectation is that local estimates in each node improve fast and converge, resulting in a limited demand for communication of estimates between nodes and reducing the processing time. We provide restricted-isometry-property (RIP)-based theoretical analysis on estimation quality. In the scenario of clean observation, it is shown that the local estimates converge to the exact sparse signal under certain technical conditions. Simulation results show that the proposed algorithms show competitive performance compared to a globally optimum distributed LASSO algorithm in the sense of convergence speed and estimation error.
\end{abstract}


%
\IEEEpeerreviewmaketitle

\section{Introduction}
Learning of sparse signals/data from a limited number of observations has become important in many applications, for example, sparse coding \cite{Elad_book_2010, Rubinstein_Elad_Dictionary_ProcIEEE_2010}, compressive sampling \cite{Donoho_2006_Compressed_sensing, CS_introduction_Candes_Wakin_2008}, dictionary learning \cite{Kreutz_Delgado_2003, Liang_distributed_DL_in_sensor_networks_2014}, etc. In the gamut of sparse learning, use of convex optimization, mainly the $\ell_1$-norm based minimization has been extensively studied \cite{Candes_Tao_universal_encoding_strategies_2006}. Other class of sparse learning algorithms are greedy methods and Bayesian approaches \cite{Chen_decentralized_bayesian_DCS_TWC_2016}. The problem of sparse signal learning becomes more challenging for a distributed setup where observations are spread across nodes of a network. This distributed sparse learning problem is relevant in applications such as big data analysis \cite{Giannakis_Big_Data_Analytics_SPMagazine_2014}, sensor networks \cite{Varshaney_OMP_joint_sparsity_pattern_recovery_TSP_2014, Varshaney_wireless_CS_distributed_sparse_TSIPN_2015} etc. In such a scenario, the sparse signal learning involves learning and exchanging information among the nodes of the network. A straightforward approach would be to process the observations at a central node. This is expected to involve sending large amounts of data over the network, resulting in a high demand in communication resource. Further, for security or privacy issues, observations and system parameters may not be accessed in a single place. With this background, we design a set of distributed sparse learning algorithms in this article. Our algorithms use convex optimization methods and do not require a large communication overhead. Each node solves a convex optimization problem to learn a local estimate. The estimates are exchanged between nodes of the network for further improvement. We expect the algorithms to converges in a limited number of iterations, thus saving communication resource and requiring a limited processing time. A low processing time is useful in low-latency applications. In this article, our main contributions are as follows:
\begin{itemize}
\item We develop $\ell_1$-norm minimization based distributed algorithms that achieve fast convergence. To include influence of estimates from neighboring nodes, we show use of several penalty functions. 
\item Incorporation of a-priori knowledge of the sparsity level is shown to achieve a faster convergence.
\item Using restricted-isometry-property based theoretical analysis, we derive bounds on signal estimation quality.
\item Simulation results confirm that the proposed algorithms provide competitive performance vis-a-vis distributed LASSO algorithm that solves a centralized problem in a distributed manner, but at the expense of slow convergence and high communication cost.
\end{itemize}
Our proposed algorithms are referred to as network basis pursuit denoising (NBPDN). The algorithms are built on the optimization formulation of much cited basis pursuit denoising (BPDN) \cite{Chen_Donoho_Saunders_1998}. We mainly modify the constrained cost function of BPDN and achieve the NBPDN.

\subsection{System Model}
\label{subsec:sys_model}
Consider a connected network with $L$ nodes. The neighborhood of node $l$ is  defined by the set, $\mathcal{N}_l \subseteq \{1,2,\hdots,L\}$. Each node is capable of receiving weighted data from other nodes in its neighborhood. The weights assigned to links between nodes can be written as a network matrix $\mbH \in \mathbb{R}^{L \times L}$ where $h_{lr}$ is the link weight from node $r$ to node $l$.
Our task is estimation of a sparse signal $\mbx$ in a distributed manner. The received signal (of size $M_l$) at node $l$ can be written as
\begin{eqnarray}
\label{eq:system_model}
\mby_l = \mbA_l \mbx + \mbe_l,
\end{eqnarray}
where $\mby_l \in \mathbb{R}^{M_l}$, $\mbA_l \in \mathbb{R}^{M_l \times N}$ is the system matrix and $\mbe_l \in \mathbb{R}^{M_l}$ is an additive noise.
\begin{assumption}
$\mbH$ is a right stochastic matrix.
\end{assumption}
The  above assumption is quite general in the sense that any non-negative network matrix can be recast as a right stochastic matrix by row normalization. 
\begin{assumption}
Observation noise is bounded, i.e., $\|\mbe_l\| \leq \epsilon$.
\end{assumption}
The above assumption is commonly used in $\ell_1$ minimization based sparse learning algorithms \cite{Candes_Romberg_Tao_2006_Stable_recovery}.
The signal of interest can be either exactly sparse ($\|\mbx\|_0 = s$) or approximately sparse ($s$ highest amplitude elements of $\mbx$ contain the maximum energy of the signal). We use $\|\cdot\|_0$, $\|\cdot\|_1$ and $\|\cdot\|$ to denote the standard $\ell_0$, $\ell_1$ and $\ell_2$ norm of an argument vector, respectively. 

\subsection{Literature Review}
\label{subsec:lit_review}
We review relevant works on sparse learning over networks in this subsection. This problem has been attempted via greedy pursuit algorithms as well as the traditional $\ell_1$ norm minimization based algorithms. In addition, various signal models have been studied for this problem, for e.g., common support (where the signal at each node is different but has the same support) \cite{Varshaney_OMP_joint_sparsity_pattern_recovery_TSP_2014} and partial support (signal support across nodes have a partial overlap) \cite{Sundman_Chatterjee_Skoglund_DistributedGreedyPursuits_2014, saikat_DIPP_TSP_2016}. Next, we discuss distributed greedy algorithms that provide computational advantage. In \cite{Sergios_greedy_sparsity_algorithm_distributed_learning_TSP_2015}, a greedy algorithm is proposed that involves exchange of observations, estimates and observation matrix to reach a consensus on the estimation over the network. Additionally, they proposed a modified algorithm that involves just the exchange of intermediate estimates. A distributed iterative hard thresholding algorithm is delveloped in \cite{Eldar_DCS_static_time_varying_network_TSP_2014} that provides sparse learning for both static and time-varying networks. An improved algorithm with lower communication cost was proposed in \cite{Eldar_mod_DIHT_ICASSP_2015}. Based on subspace pursuit \cite{Dai_2009_Subspace_pursuit} and CoSamp \cite{Needell_Tropp_2008_CoSaMP} algorithms used for centralized sparse learning, a set of distributed algorithms are proposed in \cite{Sundman_Chatterjee_Skoglund_DistributedGreedyPursuits_2014, Sundman_Chatterjee_Skoglund_2016} that provide a high computational advantage. With the same motivation of low computational complexity, a recent work on designing distributed greedy pursuit algorithms is \cite{Zaki_Venkitaraman_Chatterjee_Rasmussen_GreedySparseLearningOverNetwork_TSIPN_2017}.

Use of convex optimization for sparse learning is a much exercised area. In this area, use of $\ell_1$-norm has been investigated with considerable interest due to its optimality and robust solutions. Naturally, the strategy of solving distributed sparse learning problem using $\ell_1$ norm minimization has received more attention. A distributed approach to solve the basis pursuit denoising (BPDN) \cite{Chen_Donoho_Saunders_1998} using the method of alternating-direction-method-of-multipliers (ADMM) was proposed in \cite{Giannakis_Distributed_sparse_linear_regression_TSP_2010}. This algorithm referred to as distributed LASSO (D-LASSO), was shown to efficiently solve the distributed BPDN problem. Further work was done in \cite{Mota_Distributed_Basis_Pursuit_2012} for the noiseless setting, that means for realizing distributed basis pursuit \cite{Chen_Donoho_Saunders_1998}. Works in \cite{Duarte_Baraniuk_2005_Distributed_CS_Asilomar, Baron_DCS_2009} have proposed distributed compressive sensing algorithms using convex optimization. Other distributed algorithms use gradient search and adaptive signal processing techniques  \cite{Sergios_adaptive_algorithm_distributed_learning_TSP_2012, Li_Distributed_RLS_over_networks_TSP_2014, Sayed_sparse_distributed_learning_diffusion_TSP_2013} to solve sparse learning. These alternate approaches solve an average mean-square-error cost across the network. The local estimates are then diffused across the nodes for further refinement in the next step. Works in\cite{Eldar_DAMP_GlobalSIP_2014, Chen_decentralized_bayesian_DCS_TWC_2016} used Bayesian frameworks to design distributed sparse learning algorithms. Further, for dictionary learning which is a more general problem in sparse learning area, the works of \cite{Kreutz_Delgado_2003, Liang_distributed_DL_in_sensor_networks_2014} have investigated the problem of distributed dictionary learning using convex costs with sparsity promoting solutions. At this point we mention that the D-LASSO \cite{Giannakis_Distributed_sparse_linear_regression_TSP_2010} provides a globally optimum distributed solution using ADMM that suffers slow convergence. The slow convergence of D-LASSO is shown in \cite{Sergios_greedy_sparsity_algorithm_distributed_learning_TSP_2015}. In contrast to D-LASSO, our interest is to develop convex optimization based algorithms that are fast in convergence, albeit at the expense of global optimality. Our algorithms are locally optimum at each node of the network. This endeavor is an extension of past work in designing distributed greedy algorithms \cite{Zaki_DHTP_2017, Zaki_Venkitaraman_Chatterjee_Rasmussen_GreedySparseLearningOverNetwork_TSIPN_2017} to the regime of convex optimization.

\subsection{Notations and Preliminaries}
\label{sec:prelim}
We use calligraphic letters $\mathcal{T}$ and $\mathcal{S}$ to denote sets that are sub-sets of $\Omega \triangleq \{1, 2, \dots, N \}$. We use $|\mathcal{T}|$ and $\mathcal{T}^{c}$ to denote the cardinality and complement of the set $\mathcal{T}$, respectively.
For the matrix $\mbA \in \mathbb{R}^{M \times N}$, a sub-matrix $\mbA_{\mathcal{T}} \in \mathbb{R}^{M \times |\mathcal{T}|}$ consists of the columns of $\mbA$ indexed by $i \in \mathcal{T}$. Similarly, for $\mbx \in \mathbb{R}^{N}$, a sub-vector $\mbx_{\mathcal{T}}\in \mathbb{R}^{|\mathcal{T}|}$ is composed of the components of $\mbx$ indexed by $i \in \mathcal{T}$.
Also we denote $(\cdot)^{t}$ and $(\cdot)^{\dag}$ as transpose and pseudo-inverse, respectively. In this work $\mbA_{\mathcal{T}}^{\dag} \triangleq (\mbA_{\mathcal{T}})^{\dag}$. 
For a sparse signal $\mbx=[x_1, x_2, \hdots, x_i, \hdots, x_N]^{t}$, the support-set $\mathcal{T}$ of $\mbx$ is defined as 
$\mathcal{T} = \{ i: x_i \neq 0 \}$. 
We define a function that finds support of a vector, as follows
\begin{align} 
{\supp}(\mathbf{x}, s) & \triangleq \{ \text{the set of indices corresponding to} \nonumber \\
& \hspace{6.5mm}\text{the $s$ largest amplitude components of } \mathbf{x} \}. \nonumber 
\end{align} 
If $\mbx$ has $s$ non-zero elements then $\T = {\supp}(\mathbf{x}, s)$.
We use the standard definition of Restricted-Isometry-Property of a matrix as follows:
\begin{definition}[RIP: Restricted Isometry Property \cite{Candes_Tao_2005}]\label{def:rip} --
A matrix $\mbA \in \mathbb{R}^{M \times N}$ satisfies the RIP with Restricted Isometry Constant (RIC) $\delta_s$ if
\begin{align*}
  (1-\delta_s)\|\mbx\|^2 \leq \| \mbA \mbx \|^2 \leq (1+\delta_s) \|\mbx\|^2
\end{align*}
holds for all vectors $\mbx \in \mathbb{R}^{N}$ such that $\| \mbx \|_0 \leq s$, and $0 \leq \delta_s < 1$. The RIC satisfies the monotonicity property, i.e., $\delta_{s} \leq \delta_{2s} \leq \delta_{3s}$.
\end{definition}
We also define the restricted orthogonality constant of a matrix $\mbA \in \mathbb{R}^{M \times N}$ as follows:
\begin{definition}[ROP: Restricted Orthogonality Property\cite{Cai_Shifting_Inequality_TSP_2010}]\label{def:rop} --
For any two vectors $\mbx, \mbx' \in \mathbb{R}^{N}$ with disjoint supports such that $\|\mbx\|_0 = s, \ \|\mbx'\|_0 = s'$ and $s+s' \leq N$, the $\{s,s'\}$-\textit{restricted orthogonality constant} $\theta_{s,s'}$ (ROC), is defined as the smallest number that satisfies 
\begin{align*}
|\langle \mbA\mbx, \mbA\mbx'\rangle| \leq \theta_{s,s'} \|\mbx\| \|\mbx'\|.
\end{align*}
The ROC also satisfies the monotonicity property, i.e., $\theta_{s,s'} \leq \theta_{s_1,s_1'}$ for any $s \leq s_1$, $s' \leq s_1'$ and $s_1+s_1' \leq N$.
\end{definition}
We now state the following properties of ROC \cite{Cai_Shifting_Inequality_TSP_2010}.
\begin{proposition}
Suppose $\mbA$ has ROC $\theta_{s,s'}$ for $s+s' \leq N$. Then
 \begin{subequations}
    \begin{align}
\theta_{s,s'} \leq \delta_{s+s'}, \label{prop1:1} \\
\theta_{s,as'} \leq \sqrt{a} \theta_{s,s'}. \label{prop1:2}
\end{align}
\end{subequations}
\end{proposition}
Next we provide two lemmas to be used in deriving the results.
\begin{lemma}
\label{lem:l1_l2_bound_pruning}
Consider the standard sparse representation model $\mby=\mbA\mbx + \mbe$ with $\|\mbx\|_0 = s_1$. Let $\mathcal{S} \subseteq \{1,2,\hdots,N\}$ and $|\mathcal{S}|=s_2$. Define $\bar{\mbx}$ such that $\bar{\mbx}\leftarrow \mbA_{\mathcal{S}}^\dag \mby, \bar{\mbx}_{\mathcal{S}^c} \leftarrow \mathbf{0}$. If $\mbA$ has RIC $\delta_{s_1+s_2} < 1$, then we have the following $\ell_2$ bounds.
\begin{eqnarray*}
\|\left(\mbx-\bar{\mbx}\right)_{\mathcal{S}}\| \leq \delta_{s_1+s_2} \|\mbx-\bar{\mbx}\| + \sqrt{1+\delta_{s_2}}\|\mbe\|,
\end{eqnarray*}
and
\begin{eqnarray*}
\|\mbx-\bar{\mbx}\| \leq \sqrt{\frac{1}{1-\delta_{s_1+s_2}^2}}\|\mbx_{\mathcal{S}^c}\| + \frac{\sqrt{1+\delta_{s_2}}}{1-\delta_{s_1+s_2}}\|\mbe\|.
\end{eqnarray*}
Additionally, we have the following $\ell_1$ bounds.
\begin{eqnarray*}
\|\left(\mbx-\bar{\mbx}\right)_{\mathcal{S}}\|_1 \leq \sqrt{s_2}\delta_{s_1+s_2} \|\mbx-\bar{\mbx}\|_1 + \sqrt{s_2(1+\delta_{s_2})}\|\mbe\|,
\end{eqnarray*}
and
\begin{eqnarray*}
\|\mbx-\bar{\mbx}\|_1 \leq \sqrt{\frac{s_1+s_2}{1-\delta_{s_1+s_2}^2}}\|\mbx_{\mathcal{S}^c}\|_1 + \frac{\sqrt{(s_1+s_2)(1+\delta_{s_2})}}{1-\delta_{s_1+s_2}}\|\mbe\|.
\end{eqnarray*}
\end{lemma}
\noindent \emph{Proof:} The proof is shown in Section~\ref{sec:Theoretical_proofs}.

\begin{lemma}{\label{lem:smaller_indices_bound_l1_l2}}
Consider two vectors $\mbx$ and ${\mathbf{z}}$ with $\|\mbx\|_{0} = s_1$, $\|{\mathbf{z}}\|_{0} = s_2$ and $s_2 \geq s_1$. We have $\mathcal{S}_1 \triangleq \supp(\mbx, s_1)$ and $\mathcal{S}_2 \triangleq \supp({\mathbf{z}},s_2)$. Let $\mathcal{S}_{\nabla}$ denote the set of indices of the $s_2-s_1$ smallest magnitude elements in ${\mathbf{z}}$. Then, we have the following $\ell_2$ bound
\begin{eqnarray*}
\|\mbx_{\mathcal{S}_{\nabla}}\|  \leq \sqrt{2} \|(\mbx-{\mathbf{z}})_{\mathcal{S}_2}\| \leq \sqrt{2} \|\mbx-{\mathbf{z}}\|.
\end{eqnarray*}
Additionally, we have the following $\ell_1$ bound
\begin{eqnarray*}
\|\mbx_{\mathcal{S}_{\nabla}}\|_1  \leq  \|(\mbx-{\mathbf{z}})_{\mathcal{S}_2}\|_1 \leq \|\mbx-{\mathbf{z}}\|_1.
\end{eqnarray*}
\end{lemma}
\noindent \emph{Proof:} The proof is shown in Section~\ref{sec:Theoretical_proofs}.

The rest of the paper is organized as follows. The algorithms are proposed in Section~\ref{sec:DBP_algorithm}. The theoretical guarantees of the algorithms are discussed in Section~\ref{sec:Theoretical_analysis}. Finally, the simulation results are presented in Section~\ref{sec:simulation_results}. Section~\ref{sec:Theoretical_proofs} provides the necessary supporting lemmas and proofs of the paper.

\section{Network Basis Pursuit Denoising}
\label{sec:DBP_algorithm}
In this section, we propose two strategies of designing network basis pursuit denoising (NBPDN). For each strategy, we show use of $\ell_1$ and $\ell_2$-norm based penalties to include influence of estimations from neighboring nodes. 

\subsection{Network Basis Pursuit Denoising}
\label{subsec:NBPDN}

The pseudo-code of the NBPDN is shown in Algorithm~\ref{algo:greedy_BPDN}. In the zeroth iteration, in each node we start with the standard BPDN where $\epsilon$ is used as an error bound (as $\| \mbe_l \| \leq \epsilon$). Then, for each iteration we solve a modified cost of BPDN where we add the penalty $g(\mbx,\{\hat{\mbx}_{r,k-1}, h_{lr}\}), r \in \mathcal{N}_l$. The penalty helps to incorporate the influence of estimates from all neighboring nodes.  
\begin{algorithm}
\caption{NBPDN - Steps at Node $l$} \label{algo:greedy_BPDN}
\emph{Input}: $\mathbf{y}_l$, $\mathbf{A}_l$, $\epsilon$ \\
\emph{Initialization}:
\begin{algorithmic}
\State $k \gets 0$ \hfill ($k$ denotes iteration counter)
\end{algorithmic}
\begin{algorithmic}[1]
\State
$\hat{\mathbf{x}}_{l,0} = \hspace{-1mm} \argminl\limits_{\mathbf{x}} \,\, \|\mathbf{x}\|_{1} \,\, \text{s.t.} \,\, \|\mathbf{y}_l - \mathbf{A}_l \mathbf{x}\|_{2} \leq \epsilon $  \hfill (BPDN)
\end{algorithmic}
\emph{Iteration}:
\begin{algorithmic}
\State \textbf{repeat}
\State $k \gets k+1$ \hfill (Iteration counter)
\end{algorithmic}
\begin{algorithmic}[1]
\State $\hat{\mbx}_{l,k}  =  \argminl\limits_{\mbx}  \,\, \lambda \|\mathbf{x}\|_{1}+(1-\lambda)g(\mbx,\{\hat{\mbx}_{r,k-1}, h_{lr}\}), r \in \mathcal{N}_l  \,\,\,\,\ \text{s.t.} \, \|\mathbf{y}_l - \mathbf{A}_l \mathbf{x}\|_{2} \leq \epsilon$ 
\end{algorithmic}
\begin{algorithmic}
\State \textbf{until} \emph{stopping criterion}
\end{algorithmic}
\emph{Output}: $\hat{\mathbf{x}}_{l}$
\end{algorithm}

Examples of $g(.)$ functions that we used are as follows:
\begin{enumerate}
\item use of $\ell_1$-cost:  $g(.) = \|\mbx-\sum\limits_{r \in \mathcal{N}_l} h_{lr} \hat{\mbx}_{r,k-1}\|_1$, and
\item use of $\ell_2$-cost: $g(.) = \|\mbx-\sum\limits_{r \in \mathcal{N}_l} h_{lr} \hat{\mbx}_{r,k-1}\|$.
\end{enumerate}
In these examples, the use of $\sum\limits_{r \in \mathcal{N}_l} h_{lr} \hat{\mbx}_{r,k-1}$ is a strategy for inclusion of past estimates from neighboring nodes. We use the additive strategy for simplicity and analytical tractability. If all the solutions of neighboring nodes are sparse then the additive term is also expected to be sparse. The NBPDN that uses $\ell_1$-norm based $g(.)$ is referred to as NBPDN-1. Similarly,  the NBPDN that uses $\ell_2$-norm based $g(.)$ is referred to as NBPDN-2. A natural question arises: which norm is better to use in defining a $g(.)$ function? The use of $\ell_1$ norm promotes sparsity on the difference signal $\mbx-\sum\limits_{r \in \mathcal{N}_l} h_{lr} \hat{\mbx}_{r,k-1}$. Therefore, our hypothesis is that the $\ell_1$ norm based $g(.)$ function promotes a sparse solution for $\mbx$. The solution is supposed to have a high overlap between its support and the support of the sparse signal $\sum\limits_{r \in \mathcal{N}_l} h_{lr} \hat{\mbx}_{r,k-1}$. In the cost minimization, the parameter $\lambda$ needs to carefully chosen to keep a balance between the sparsity promoting function $\| \mbx \|_1$ and the $g(.)$ function. Next, we mention our main theoretical result on bounding the estimation error. 

\emph{Main Theoretical Result:} Over iterations, the NBPDN-1 algorithm has a recurrence relation given by,
\begin{eqnarray*}
\|\mbx - \hat{\mbx}_{l,k}\| \leq \sum\limits_{l=1}^{L} \varrho_1 \|\mbx-\hat{\mbx}_{l,k-1}\| + \upsilon_1 \|\mbx_{\T^c}\|_1 + \zeta_1 \epsilon,
\end{eqnarray*}
where $\varrho_1, \upsilon_1$ and $\zeta_1$ are deterministic constants depending on network weights $h_{lr}$, and RIP constant $\delta$ and ROC constant $\theta$ of $\mbA_l$. Also, for a fixed $s$, $\T$ is the $s$-support set of the signal $\mbx$. The recurrence relation bounds the estimation error at node $l$ in iteration $k$.  The above recurrence relation is further used to show that the estimation error is bounded as
\begin{eqnarray}
\|\mbx-\hat{\mbx}_l\| \leq \upsilon_2 \|\mbx_{\T^c}\|_1 + \zeta_2 \epsilon.
\end{eqnarray}
The above bound holds at each node of the network if RIC and ROC constants of the local observation matrix $\mbA_l$ satisfy certain conditions. Under the assumption of no observation noise and the signal being exactly $s$-sparse, the NBPDN achieves exact estimate of $\mbx$ at every node when $\delta_{2s} < 0.472$. This result holds irrespective of the use of $\ell_1$ or $\ell_2$-norm  in $g(.)$ function. We will investigate the main result in detail later in section~\ref{sec:Theoretical_analysis}.

\subsection{Pruning based Network Basis Pursuit Denoising}

In this subsection, we develop a derivative of NBPDN with a-priori information of sparsity level $\| \mbx \|_0 =s$. This new derivative is referred to as pruned NBPDN (pNBPDN), shown in Algorithm~\ref{algo:greedy_s_prune_BPDN}. The use of sparsity level brings a flavor for comparative studies with greedy algorithms where sparsity level is typically used as a-priori information. In Algorithm~\ref{algo:greedy_s_prune_BPDN}, we apply pruning to sparsify the estimates $\tilde{\mbx}_{l,k}$. We use the support-finding function $\supp(.)$ to determine the indices of the highest $s$-amplitudes in the estimate (see Step 2 of the algorithm). We then do a projection operation to get an $s$-sparse estimate $\hat{\mbx}_{l,k}$ (see Step 3 of the algorithm). We again consider two examples of the $g(.)$ function. The pNBPDN that uses $\ell_1$-norm based $g(.)$ is referred to as pNBPDN-1. Similarly,  the NBPDN that uses $\ell_2$-norm based $g(.)$ is referred to as pNBPDN-2. 

\begin{algorithm}
\caption{Pruned NBPDN - Steps at Node $l$} \label{algo:greedy_s_prune_BPDN}
\emph{Input}: $\mathbf{y}_l$, $\mathbf{A}_l$, $s$, \ $\epsilon$ \\
\emph{Initialization}:
\begin{algorithmic}
\State $k \gets 0$ \hfill ($k$ denotes iteration counter)
\end{algorithmic}
\begin{algorithmic}[1]
\State $\tilde{\mathbf{x}}_{l,0} = \hspace{-1mm} \argminl\limits_{\mathbf{x}} \|\mathbf{x}\|_{1} \,\, \text{s.t.} \,\, \|\mathbf{y}_l - \mathbf{A}_l \mathbf{x}\|_{2} \leq \epsilon $  \hfill (BPDN)
\State $\hat{\T}_{l,0} \gets \supp (\tilde{\mbx}_{l,0},s)$
\State $\hat{\mbx}_{l,0} \,\,\ \text{such that} \,\,\ \hat{\mbx}_{\hat{\T}_{l,0}} \gets \mbA^{\dag}_{l,\hat{\T}_{l,0}} \mby_l \ ; \,\ \hat{\mbx}_{\hat{\T}^{c}_{l,0}} \gets \mathbf{0}$ \hfill (Pruning)
\end{algorithmic}
\emph{Iteration}:
\begin{algorithmic}
\State \textbf{repeat}
\State $k \gets k+1$ \hfill (Iteration counter)
\end{algorithmic}
\begin{algorithmic}[1]
\State $\tilde{\mbx}_{l,k}  =  \argminl\limits_{\mbx}  \lambda \|\mathbf{x}\|_{1}+(1-\lambda)g(\mbx,\{\hat{\mbx}_{r,k-1}, h_{lr}\}), r \in \mathcal{N}_l  \,\,\,\,\ \text{s.t.} \, \|\mathbf{y}_l - \mathbf{A}_l \mathbf{x}\|_{2} \leq \epsilon$ \hfill (Adapt)
\State $\hat{\T}_{l,k} \gets \supp (\tilde{\mbx}_{l,k},s)$
\State $\hat{\mbx}_{l,k} \,\,\ \text{such that} \,\,\ \hat{\mbx}_{\hat{\T}_{l,k}} \gets \mbA^{\dag}_{l,\hat{\T}_{l,k}} \mby_l \ ; \,\ \hat{\mbx}_{\hat{\T}^{c}_{l,k}} \gets \mathbf{0}$ \hfill (Pruning)
\end{algorithmic}
\begin{algorithmic}
\State \textbf{until} \emph{stopping criterion}
\end{algorithmic}
\emph{Output}: $\hat{\mathbf{x}}_{l}, \ \mathbf{r}_{l}$
\end{algorithm}

\emph{Main Theoretical Result:} For pNBPDN, the estimation error is bounded as
\begin{eqnarray}
\|\mbx-\hat{\mbx}_l\| \leq \text{constant} \times \epsilon,
\end{eqnarray}
The above bound holds at each node in the network and requires certain conditions on the RIC and ROC constants of the local observation matrix $\mbA_l$. Under the assumption of no observation noise and the signal being exactly sparse, the pNBPDN achieves exact estimate of $\mbx$ at every node when $\delta_{2s} < 0.472$.

\subsection{Discussions}
For the scenario of no cooperation over network, that is, if $\mbH$ is an identity matrix, NBPDN-1 is same as the BPDN algorithm. Interestingly, it has been shown that the BPDN has an RIP condition that $\delta_{2s}(\mbA_l) < 0.472$ \cite{Cai_Shifting_Inequality_TSP_2010} for bounded reconstruction. Therefore, according to our analysis the RIP conditions for BPDN and NBPDN are comparable. The use of a-priori knowledge in pruned NBPDN does not change the RIP condition. The pruned BPDN (pNBPDN) is more close to distributed greedy algorithms that work with the same system setup. Examples of the greedy algorithms are network greedy pursuit (NGP) \cite{Zaki_Venkitaraman_Chatterjee_Rasmussen_GreedySparseLearningOverNetwork_TSIPN_2017} and distributed hard thresholding pursuit (DHTP) \cite{Zaki_DHTP_2017}. We mention that NGP and DHTP have RIP conditions $\delta_{3s}(\mbA_l) < 0.362$ and $\delta_{3s}(\mbA_l) < 0.333$, respectively. It can be seen that RIP conditions for these two distributed greedy algorithms are more strict compared to the proposed NBPDN.


\section{Theoretical Analysis}
\label{sec:Theoretical_analysis}

In this section we discuss the main theoretical results of this article. For notational clarity, we use RIC constant $\delta_{s_1} \triangleq \underset{l}{\max}\{\delta_{s_1}\left(\mbA_l\right)\}$ and ROC constant $\theta_{s_1,s_2} \triangleq \underset{l}{\max}\{\theta_{s_1,s_2}\left(\mbA_l\right)\}$ where $s_1$, $s_2$ are constants. We first start with the analysis of NBPDN. Let us define $\mbz_{l,k} \triangleq \hat{\mbx}_{l,k} - \mbx$. In NBPDN,  $\|\mbA \hat{\mbx}_{l,k} -\mby\| \leq \epsilon$. Also, we have the model constraint $\|\mby_l-\mbA_l\mbx\| \leq \epsilon$. Then we have
\begin{eqnarray}
\label{eq:h_bound_eq_1}
\begin{array}{rl}
\|\mbA_l \mbz_{l,k} \| & = \|\mbA_l(\hat{\mbx}_{l,k}-\mbx)\| \\ & \stackrel{(a)}{\leq} \|\mbA_l \hat{\mbx}_{l,k} -\mby_l\| + \|\mby_l-\mbA_l\mbx\| \leq 2\epsilon,
\end{array}
\end{eqnarray}
where $(a)$ follows from the triangle inequality. We can now partition $\mbz_{l,k}$ into disjoint sparse vectors such that, $\mbz_{l,k} = \left\{ \bigcup\limits_{j \geq 0} \mbz_{\T_j} \right\} \cup \mbz_{\T_{*}}$. We have dropped the subscript $\{l,k\}$ in individual partitioned vectors for notational clarity. In the above partition, $\T_0 \triangleq \supp(\mbx,s)$, $\T_{*} \triangleq \supp(\mbz_{\T_0^c},a)$,  $\T_{1} \triangleq \supp(\mbz_{(\T_0 \cup \T_{*})^c},b)$, $\T_{2} \triangleq \supp(\mbz_{(\T_0 \cup \T_1 \cup \T_{*})^c},b)$ and so on. Here, we assume that $a$ and $b$ are two positive integers satisfying the condition $a < b \leq 4a$. We now state a derivative of the shifting inequality from \cite{Cai_Shifting_Inequality_TSP_2010}.
\begin{lemma}[Consequence of the shifting inequality \cite{Cai_Shifting_Inequality_TSP_2010}]
\label{lem:shifting_inequality_lemma}
\begin{eqnarray}
\sum\limits_{i \geq 1} \|\left(\mbz_{l,k}\right)_{\T_i}\| \leq \frac{\|\left(\mbz_{l,k}\right)_{\T_0^c}\|_1}{\sqrt{b}}
\end{eqnarray}
\end{lemma}
The above lemma follows from applying the shifting inequality lemma \cite[Lemma 3]{Cai_Shifting_Inequality_TSP_2010} iteratively to the partitioning of $\|\mbz_{l,k}\|$.

\subsection{Bounds on estimation error for NBPDN-1}
For NBPDN-1 algorithm, $g(\mbx,\{\hat{\mbx}_{r,k-1}, h_{lr}\}_{r \in \mathcal{N}_l}) = \|\mbx-\sum_{r \in \mathcal{N}_l} h_{lr} \hat{\mbx}_{r,k-1}\|_1$.
We first present the following lemma which will be used later to prove the results.
\begin{lemma}
\label{lem:inequality_lemma}
For $\lambda > 1/2$,
\begin{subequations}
    \begin{align}
& \|\left(\mbz_{l,k}\right)_{\T_0^c}\|_1 \leq \frac{1}{2 \lambda -1}\|\left(\mbz_{l,k}\right)_{\T_0}\|_1 + \frac{2\lambda}{2\lambda -1}\|\mbx_{\T_0^c}\|_1. \label{eq:lemma_algo1_1} \\
& \text{Alternately, for any } \lambda, \nonumber \\
& \begin{array}{rl}
\|\left(\mbz_{l,k}\right)_{\T_0^c}\|_1 \leq & (2 \lambda -1)\|\left(\mbz_{l,k}\right)_{\T_0}\|_1 + 2\lambda\|\mbx_{\T_0^c}\|_1  \\
& + 2(1-\lambda)\|\mbx  - \acute{\mbx}_{l,k}\|_1. \label{eq:lemma_algo1_2}
\end{array}
\end{align}
\end{subequations}
\end{lemma}
Here we define
\begin{eqnarray*}
 \acute{\mbx}_{l,k} \triangleq \sum\limits_{r \in \mathcal{N}_l} h_{lr} \hat{\mbx}_{r,k-1}.
\end{eqnarray*}
The proof of Lemma~\ref{lem:inequality_lemma} is shown in Section~\ref{sec:Theoretical_proofs}. Next, we use the results of Lemma~\ref{lem:inequality_lemma} to provide a bound on the estimation error of NBPDN-1 in the following theorem.
\begin{theorem}[Bound on estimation error]
\label{thm:l1_conv_theorem1}
Let $\delta_{s+a}+\sqrt{\frac{s}{b}}\frac{\theta_{s+a,b}}{2\lambda -1} < 1$ be satisfied for $\lambda > 1/2$. Then the estimation error at iteration $k$ is bounded as
\begin{eqnarray*}
\begin{array}{l}
\|\mbx-\hat{\mbx}_{l,k}\| \\ \hspace{5mm}\leq \frac{4 \lambda \sqrt{s(1+\delta_{s+a})}}{c_1(2\lambda -1)} \epsilon + \frac{2\lambda}{2\lambda -1}\left[1 + \frac{2\lambda}{2\lambda -1} \sqrt{\frac{s}{b}} \frac{\theta_{s+a,b}}{c_1}\right]\|\mbx_{\T_0^c}\|_1,
\end{array}
\end{eqnarray*}
where $c_1 = 1-\delta_{s+a}-\sqrt{\frac{s}{b}}\frac{\theta_{s+a,b}}{2\lambda -1} > 0$.
\end{theorem}
Proof of the above theorem is shown in Section~\ref{sec:Theoretical_proofs}. The above result shows that estimation error is bounded by noise parameter $\epsilon$ and signal parameter $\|\mbx_{\T_0^c}\|_1$. In case of no observation noise ($\epsilon=0$) and for $s$-sparse signal ($\|\mbx_{\T_0^c}\|_1=0$), we have perfect estimation. We now discuss the required conditions for the above theorem to be valid. On condition is that the observation matrix $\mbA_l$ should satisfy $\delta_{s+a}+\sqrt{\frac{s}{b}}\frac{\theta_{s+a,b}}{2\lambda -1} < 1$. Under the assumption of $s=4a=b$, the above requirement can be upper bounded as 
\begin{eqnarray*}
\begin{array}{l}
\delta_{1.25s}+\frac{\theta_{1.25s,s}}{2\lambda -1} \leq \delta_{2s}+\frac{\sqrt{1.25}}{2\lambda-1} \delta_{2s},
\end{array}
\end{eqnarray*} 
where we have used \eqref{prop1:1} and \eqref{prop1:2} to simplify the expression. The requirement can now be written as $\delta_{2s} < \frac{2\lambda-1}{2\lambda+0.12}$. It can be seen that the RIC constant $\delta_{2s}$ is a function of $\lambda$ and becomes stricter with decreasing $\lambda$. The most relaxed condition of $\delta_{2s} < 0.472 $ is achieved for $\lambda$ close to one. In addition, the above result is valid only for $\lambda > 1/2$. We next show an estimation error bound for a general $\lambda$. This requires the following recurrence inequality result over iterations.
\begin{theorem}[Recurrence inequality]
\label{thm:l1_recurrence_theorem2}
Under the condition $\delta_{s+a}+\sqrt{\frac{s}{b}} (2\lambda'  -1)\theta_{s+a,b} < 1$, at iteration $k$, we have
\begin{eqnarray*}
\|\mbz_{l,k}\| _1\leq c_2 \sum\limits_{r \in \mathcal{N}_l} h_{lr} \|\mbz_{r,k-1}\|_1+ c_3 \|\mbx_{\T_0^c}\|_1 + c_4 \epsilon,
\end{eqnarray*}
\begin{eqnarray*}
\begin{array}{l}
\hspace{-3mm}\text{where} \,\,\ \lambda' = \max\{\lambda,1/2\}, c_2 = 2(1-\lambda) \left[1 + 2\lambda \sqrt{\frac{s}{b}} \frac{\theta_{s+a,b}}{c_5}\right], \\
c_3 = 2\lambda \left[1 + 2\lambda \sqrt{\frac{s}{b}} \frac{\theta_{s+a,b}}{c_5}\right], c_4 = \frac{4 \lambda \sqrt{s(1+\delta_{s+a})}}{c_5}, \\
\text{and} \,\ c_5 = 1-\delta_{s+a}-\sqrt{\frac{s}{b}} (2\lambda' -1)\theta_{s+a,b}.
\end{array}
\end{eqnarray*}
\end{theorem}
With the help of the recurrence inequality, we show the following result for a general $\lambda$.
\begin{theorem}[Bound on estimation error]
\label{thm:l1_convergence_theorem2}
If $\delta_{s+a}+\sqrt{\frac{s}{b}}\theta_{s+a,b} < 1$, then at iteration $k$, the estimation error is bounded by
\begin{eqnarray*}
\|\mbx-\hat{\mbx}_{l,k}\| \leq d_1 \epsilon + d_2 \|\mbx_{\T_0^c}\|_1,
\end{eqnarray*}
where $d_1 = \frac{c_2^k-1}{c_2-1}(c_4+c_2c_7), d_2 = \frac{c_2^k-1}{c_2-1}(c_3+c_2c_6), c_6 = 2 \left[1 + 2 \sqrt{\frac{s}{b}} \frac{\theta_{s+a,b}}{c_5}\right],$ and $c_7 = \frac{4 \sqrt{s(1+\delta_{s+a})}}{c_5}$.
\end{theorem}
The detailed proofs of the above two theorems is given in Section~\ref{sec:Theoretical_proofs}. It follows that, for the case of no observation noise ($\epsilon=0$) and $s$-sparse signal ($\|\mbx_{\T_0^c}\|_1=0$), we have perfect estimation. The above two theorems are valid for two different conditions on the RIC and ROC constants of $\mbA_l$. We can easily see that with the restriction $\lambda' = \max\{\lambda,1/2\}$, the second condition of $\delta_{s+a}+\sqrt{\frac{s}{b}}\theta_{s+a,b} < 1$ is stricter compared to the condition in Theorem~\ref{thm:l1_recurrence_theorem2}. Hence, both the above theorems are valid under the condition $\delta_{s+a}+\sqrt{\frac{s}{b}}\theta_{s+a,b} < 1$. Following the earlier discussion after Theorem~\ref{thm:l1_conv_theorem1}, the above condition can be bounded as $\delta_{2s} < 0.472$ for $s = 4a = b$. This is the requirement on the RIC constant that we had earlier stated in the main results.

The two bounding results in Theorem~\ref{thm:l1_conv_theorem1} and Theorem~\ref{thm:l1_convergence_theorem2} provide different quality of bounds on the estimation error. The bound in Theorem~\ref{thm:l1_conv_theorem1} remains same at any iteration $k$, i.e., the estimation error is upper bounded by a constant value. But this bounding comes with a compromise on the RIC constant $\delta_{2s}$ which can take the value $(0,0.472)$ depending on $\lambda$. Additionally, this bound requires the condition $\lambda > 1/2$. On the other hand, Theorem~\ref{thm:l1_convergence_theorem2} provides bound on the estimation error for any $\lambda$. Also, the requirement on the RIC constant $\delta_{2s}<0.472$ does not depend on the value of $\lambda$. The only relaxation is in the actual value of the bound that is dependent on the iteration value $k$. This is because our recurrence inequality is an upper bound and increases with increase in the number of iterations. Next, we show the results for NBPDN-2.

\subsection{Bounds on estimation error for NBPDN-2}
For NBPDN-2 algorithm, $g(\mbx,\{\hat{\mbx}_{r,k-1}, h_{lr}\}_{r \in \mathcal{N}_l}) = \|\mbx-\sum_{r \in \mathcal{N}_l} h_{lr} \hat{\mbx}_{r,k-1}\|$.
We first derive the following bounds on the result of the minimization function. 
\begin{lemma}
\label{lem:inequality_lemma_algo5}
For $\lambda > 1/2$,
\begin{subequations}
    \begin{align}
& \|\left(\mbz_{l,k}\right)_{\T_0^c}\|_1 \leq \frac{1}{2 \lambda -1}\|\left(\mbz_{l,k}\right)_{\T_0}\|_1 + \frac{2\lambda}{2\lambda -1}\|\mbx_{\T_0^c}\|_1. \label{eq:lemma_algo5_1} \\
& \text{Alternately, for any } \lambda, \nonumber \\
& \begin{array}{rl}
\|\left(\mbz_{l,k}\right)_{\T_0^c}\|_1 \leq & \left[1-\frac{1-\lambda}{\lambda \sqrt{s}}\right]\|\left(\mbz_{l,k}\right)_{\T_0}\|_1 + 2\|\mbx_{\T_0^c}\|_1  \\
& + \frac{2(1-\lambda)}{\lambda}\|\mbx  - \acute{\mbx}_{l,k}\|.\label{eq:lemma_algo5_2}
\end{array}
\end{align}
\end{subequations}
\end{lemma}
The proof is shown in Section~\ref{sec:Theoretical_proofs}. We next state a bound on the estimation error for the initialization step which will be used to prove the overall bound later.
\begin{lemma}
\label{lem:BPDN_convergence_result}
For the initialization step, we have 
\begin{eqnarray*}
\|\mbx-\hat{\mbx}_{l,0}\| \leq c_{12} \|\mbx_{\T_0^c}\|_1 + c_{13} \epsilon,
\end{eqnarray*}
where $c_{12} = \frac{2}{\sqrt{b}} \left[1 + \frac{\theta_{s+a,b}\sqrt{1+s/b}}{1-\delta_{s+a}-\sqrt{\frac{s}{b}}\theta_{s+a,b}}\right], c_{13} = 2 \sqrt{1+\frac{s}{b}} \frac{\sqrt{1+\delta_{s+a}}}{1-\delta_{s+a}-\sqrt{\frac{s}{b}}\theta_{s+a,b}} $.
\end{lemma}
The above lemma follows from \cite[Theorem 4]{Cai_Shifting_Inequality_TSP_2010} as BPDN algorithm is used in the initialization step. Next, we show the convergence of Algorithm 2 in the following theorem.
\begin{theorem}[Bound on estimation error]
\label{thm:l2_conv_theorem1}
Let $\delta_{s+a}+\sqrt{\frac{s}{b}}\frac{\theta_{s+a,b}}{2\lambda -1} < 1$ be satisfied for $\lambda > 1/2$. Then the estimate at any iteration $k$, $\hat{\mbx}_{l,k}$ is bounded as
\begin{eqnarray*}
\begin{array}{l}
\|\mbx-\hat{\mbx}_{l,k}\| \\ \hspace{5mm}\leq \frac{4 \lambda \sqrt{s(1+\delta_{s+a})}}{c_1(2\lambda -1)} \epsilon + \frac{2\lambda}{2\lambda -1}\left[1 + \frac{2\lambda}{2\lambda -1} \sqrt{\frac{s}{b}} \frac{\theta_{s+a,b}}{c_1}\right]\|\mbx_{\T_0^c}\|_1.
\end{array}
\end{eqnarray*}
\end{theorem}
The detailed proof of the above theorem is given in Section~\ref{sec:Theoretical_proofs}. It is interesting to see the similarity between Theorem~\ref{thm:l1_conv_theorem1} and Theorem~\ref{thm:l2_conv_theorem1}. It is easy to see that for $s =4a=b$, the condition on RIC constant is dependent on $\lambda$ and can be written as $\delta_{2s} < \frac{2\lambda-1}{2\lambda+0.12}$. The range of the RIC constant is given by $\delta_{2s} \in (0,0.472)$ for $\lambda \in (0.5,1]$. We next bound the estimation error for NBPDN-2 for a general $\lambda$. We first present the following recurrence relation at iteration $k$.
\begin{theorem}[Recurrence inequality]
\label{thm:l2_recurrence_theorem2}
Under the condition $\delta_{s+a}+\sqrt{\frac{s}{b}} (2\lambda^{\prime \prime} -1)\theta_{s+a,b} < 1$, at iteration $k$, we have
\begin{eqnarray*}
\|\mbz_{l,k}\| \leq c_9 \sum\limits_{r \in \mathcal{N}_l} h_{lr} \|\mbz_{r,k-1}\|+ c_{10} \|\mbx_{T_0^c}\|_1 + c_{11} \epsilon,
\end{eqnarray*}
where 
\begin{eqnarray*}
\begin{array}{l}
\lambda^{\prime \prime} = \max\left\{\lambda,\frac{1}{1+\sqrt{s}}\right\}, 
c_8 = 1-\delta_{s+a}-\frac{\lambda^{\prime \prime}(1+\sqrt{s})-1}{\sqrt{b}}\theta_{s+a,b}, \\
c_9 = \frac{2(1-\lambda)}{\lambda \sqrt{b}} \left[1 + \left(1+\frac{\lambda^{\prime \prime}(1+\sqrt{s})-1}{\sqrt{b}}\right)\frac{\theta_{s+a,b}}{c_8}\right], \\
c_{10} = \frac{2}{\sqrt{b}} \left[1 + \left(1+\frac{\lambda^{\prime \prime}(1+\sqrt{s})-1}{\sqrt{b}}\right)\frac{\theta_{s+a,b}}{c_8} \right] \\
\text{and}  \,\ c_{11} = \left(1+\frac{\lambda^{\prime \prime}(1+\sqrt{s})-1}{\sqrt{b}}\right)\frac{2 \sqrt{1+\delta_{s+a}}}{c_8}.
\end{array}
\end{eqnarray*}
\end{theorem}
The above recurrence relation can be used to derive the following bound on the estimation error.
\begin{theorem}[Bound on estimation error]
\label{thm:l2_convergence_theorem2}
If $\delta_{s+a}+\sqrt{\frac{s}{b}}\theta_{s+a,b} < 1$, then at iteration $k$, the estimate $\hat{\mbx}_{l,k}$ is bounded by
\begin{eqnarray*}
\|\mbx-\hat{\mbx}_{l,k}\| \leq d_3 \epsilon + d_4 \|\mbx_{\T_0^c}\|_1,
\end{eqnarray*}
where $d_3 = \frac{c_9^k-1}{c_9-1}(c_{11}+c_9 c_{13})$, and $d_4 = \frac{c_9^k-1}{c_9-1}(c_{10}+c_9 c_{12})$.
\end{theorem}
The detailed proofs of the above two theorems is given in Section~\ref{sec:Theoretical_proofs}. For $s=4a=b$, the RIC requirement reduces to $\delta_{2s} < 0.472$. The trade-off between quality of the estimation error bound and the RIC constant $\delta_{2s}$ is similar to that discussed for NBPDN-1. That is, Theorem~\ref{thm:l2_conv_theorem1} provides a better bound at the cost of stricter $\delta_{2s}$ and for $\lambda > 1/2$. On the other hand, Theorem~\ref{thm:l2_convergence_theorem2} gives a looser bound which is valid for any $\lambda$ and RIC constant $\delta_{2s} < 0.472$. It even follows that for the case of no observation noise ($\epsilon=0$) and for $s$-sparse signal ($\|\mbx_{\T_0^c}\|_1=0$), we have perfect estimation for NBPDN-2.

\subsection{Bounds on estimation error for pNBPDN-1 and 2}
In this section, we derive bounds on the estimation error of pNBPDN-1 and pNBPDN-2. Recall that for the pruned algorithms, we assumed that the signal is $s$-sparse, i.e., $\| \mbx \|_0 =s$. It can be seen that Algorithm pNBPDN-1 is similar to Algorithm NBPDN-1 except the additional use of pruning step. We will use the following lemmas to bound the error in the pruning step.

\begin{lemma}
\label{lem:pruning_bound_lemma1}
Assume $\|\mbx\|_0 = s$, then we have
\begin{eqnarray*}
\|\mbx-\hat{\mbx}_{l,k}\| \leq \sqrt{\frac{2}{1-\delta_{2s}^2}}\|\mbx-\tilde{\mbx}_{l,k}\| + \frac{\sqrt{1+\delta_{s}}}{1-\delta_{2s}}\epsilon.
\end{eqnarray*}
Additionally, the following $\ell_1$ bound holds true.
\begin{eqnarray*}
\|\mbx-\hat{\mbx}_{l,k}\|_1 \leq \sqrt{\frac{2s}{1-\delta_{2s}^2}}\|\mbx-\tilde{\mbx}_{l,k}\|_1 + \frac{\sqrt{2s(1+\delta_{s})}}{1-\delta_{2s}}\epsilon.
\end{eqnarray*}
\end{lemma}
The proof is shown in Section~\ref{sec:Theoretical_proofs}. Now, we can extend Theorem~\ref{thm:l1_conv_theorem1} and Theorem~\ref{thm:l1_convergence_theorem2} to get the following result.
\begin{theorem}[Bound on estimation error]
\label{thm:l1_pruning_theorem_convergence}
For pNBDN-1algorithm, we have the following bounds on the estimation error.
\begin{enumerate}
\item Let $\delta_{s+a}+\sqrt{\frac{s}{b}}\frac{\theta_{s+a,b}}{2\lambda -1} < 1$ be satisfied for $\lambda > 1/2$. Then the estimation error at any iteration $k$ is bounded as
\begin{eqnarray*}
\begin{array}{l}
\hspace{0mm}\|\mbx-\hat{\mbx}_{l,k}\| \leq \left(\frac{4\lambda \sqrt{2s(1+\delta_{s+a})}}{c_1(2\lambda -1)\sqrt{1-\delta_{2s}^2}} + \frac{\sqrt{1+\delta_s}}{1-\delta_{2s}}\right) \epsilon.
\end{array}
\end{eqnarray*}
\item If $\delta_{s+a}+\sqrt{\frac{s}{b}}\theta_{s+a,b} < 1$, then at iteration $k$, the estimation error is bounded by
\begin{eqnarray*}
\begin{array}{l}
\hspace{-4mm}\|\mbx-\hat{\mbx}_{l,k}\| \\ \leq \frac{c_{14}^k-1}{c_{14}-1}\left((c_{4}+c_{14}c_{7}) \sqrt{\frac{2s}{1-\delta_{2s}^2}} + \frac{(1+c_{14})\sqrt{2s(1+\delta_{s})}}{1-\delta_{2s}}\right) \epsilon,
\end{array}
\end{eqnarray*}
where $c_{14} = c_2\sqrt{\frac{2s}{1-\delta_{2s}^2}}$. 
\end{enumerate}
\end{theorem}
The detailed proof of the above theorem is given in Section~\ref{sec:Theoretical_proofs}. Following similar arguments in the discussion of NBPDN-1, it follows that the first bound on estimation error is valid for $\delta_{2s} < \frac{2\lambda-1}{2\lambda+0.12}$ for $s = 4a = b$. Interestingly, the pruning step of pNBDN-1 does not change the RIC constant requirement. The second bound requires an RIC constant of $\delta_{2s} < 0.472$. It also follows that pNBPDN-1 achieves perfect reconstruction under the scenario of no observation noise.

Finally we consider pNBPDN-2. The pNBPDN-2 algorithm is same as the NBPDN-2 with the additional pruning step. Hence, we can extend Theorem~\ref{thm:l2_conv_theorem1} and Theorem~\ref{thm:l2_convergence_theorem2} to get the following results.
\begin{theorem}[Bound on estimation error]
\label{thm:l2_pruning_theorem_convergence}
For pNBDN-2 algorithm, we have the following bounds on the estimation error.
\begin{enumerate}
\item Let $\delta_{s+a}+\sqrt{\frac{s}{b}}\frac{\theta_{s+a,b}}{2\lambda -1} < 1$ be satisfied for $\lambda > 1/2$. Then the estimation error at any iteration $k$ is bounded as
\begin{eqnarray*}
\begin{array}{l}
\hspace{-4mm}\|\mbx-\hat{\mbx}_{l,k}\| \leq \left(\frac{4\lambda \sqrt{2s(1+\delta_{s+a})}}{c_1(2\lambda -1)\sqrt{1-\delta_{2s}^2}} + \frac{\sqrt{1+\delta_s}}{1-\delta_{2s}}\right) \epsilon.
\end{array}
\end{eqnarray*}
\item If $\delta_{s+a}+\sqrt{\frac{s}{b}}\theta_{s+a,b} < 1$, then at iteration $k$, the estimation error is bounded as
\begin{eqnarray*}
\begin{array}{l}
\hspace{-2mm}\|\mbx-\hat{\mbx}_{l,k}\| \\ \leq \frac{c_{15}^k-1}{c_{15}-1}\left((c_{11}+c_{15}c_{13}) \sqrt{\frac{2}{1-\delta_{2s}^2}} + (1+c_{15})\frac{\sqrt{1+\delta_{s}}}{1-\delta_{2s}}\right) \epsilon,
\end{array}
\end{eqnarray*}
where $c_{15} = c_9\sqrt{\frac{2}{1-\delta_{2s}^2}}$. 
\end{enumerate}
\end{theorem}
The detailed proof of the above theorem is given in Section~\ref{sec:Theoretical_proofs}. Under the assumption of $s=4a=b$, it can be seen that the requirement on the RIC constant for the first and second bound reduces to $\delta_{2s} < \frac{2\lambda-1}{2\lambda+0.12}$ and $\delta_{2s} < 0.472$ respectively. Additionally, in the absence of observation noise, pNBPDN-2 achieves perfect reconstruction.

\section{Simulation Results}
\label{sec:simulation_results}
In this section, we study the performance of the proposed algorithms using simulations. We first describe the simulation setup and the test scenarios. Simulation results are then discussed.

\subsection{Simulation Setup}
We consider a randomly chosen connected network with $L$ nodes where each node is connected to other $d$ nodes. The parameter $d$ is referred to as the `degree of the network' that gives a measure of the network connection density. We have $d = |\mathcal{N}_l|$. Given the edge matrix $\mathcal{E}$ of the network, we can generate a right stochastic network matrix $\mbH$. This can be done by ensuring that the following conditions are satisfied,
\begin{eqnarray*}
\mbH \mathbf{1} = \mathbf{1}, \, \text{and} \,\ \forall(i,j), h_{ij} \geq 0, \forall(i,j) \notin \mathcal{E}, h_{ij} = 0.
\end{eqnarray*}
We use the system model \eqref{eq:system_model} to generate different realizations of $\mby_l$ by randomly generating  sparse signal $\mbx$ and observation noise $\mbe_l$ for given observation matrices $\mbA_l$. The observation noise is i.i.d. Gaussian. We use Gaussian sparse signals, i.e., sparse signals with non-zero elements chosen from a Gaussian distribution \cite{Chatterjee_Sundman_Vehkapera_Skoglund_TSP_2012}. To average out results for $\mbA_l$, we also generate many instances of $\mbA_l$ and perform Monte Carlo simulations. The performance metric used to compare the various algorithms is the mean signal-to-estimation-noise-ratio (mSENR),
\begin{equation*}
\text{mSENR} = \frac{1}{L} \sum\limits_{l=1}^{L} \frac{\mathbb{E}\{\|\mbx\|^{2}\}}{\mathbb{E}\{\|\mbx -\hat{\mbx}_l\|^{2}\}},
\end{equation*}
where $\mathbb{E}(.)$ is the sampling average of the simulation. We define the signal-to-noise ratio (SNR) for node $l$ as
\begin{equation*}
\text{SNR}_l =  \frac{\mathbb{E}\{\|\mbx\|^{2}\}}{\mathbb{E}\{\|\mbe_l\|^{2}\}}.
\end{equation*}
For simplicity, we assume that SNR is same at all the nodes, that is $\forall l, \text{SNR}_l = \text{SNR}$. Also, we consider the observation matrices to be of the same size, i.e., $\forall l, M_l = M$. We simulate D-LASSO, NBPDN-1, NBPDN-2, pNBPDN-1 and pNBPDN-2 algorithms. The reason to compare with D-LASSO is that it provides a benchmark performance -- centralized solution of BPDN in a distributed manner. For all the above algorithms the stopping criterion is assumed to be a corresponding maximum number of iterations allowed. The choice of maximum number of iterations differs for various experiments. This allows us to enforce a measure of communication constraints and/or processing time constraints. For all the experiments below, we set observation size $M = 100$ and signal dimension $N = 500$ with sparsity level $s = 20$. The network is assumed to have 20 nodes ($L =20$) with degree, $d = 4$. 

\subsection{Experiment on Convergence Speed}
\begin{figure*}
\hspace{-14mm}
\includegraphics[scale=0.49]{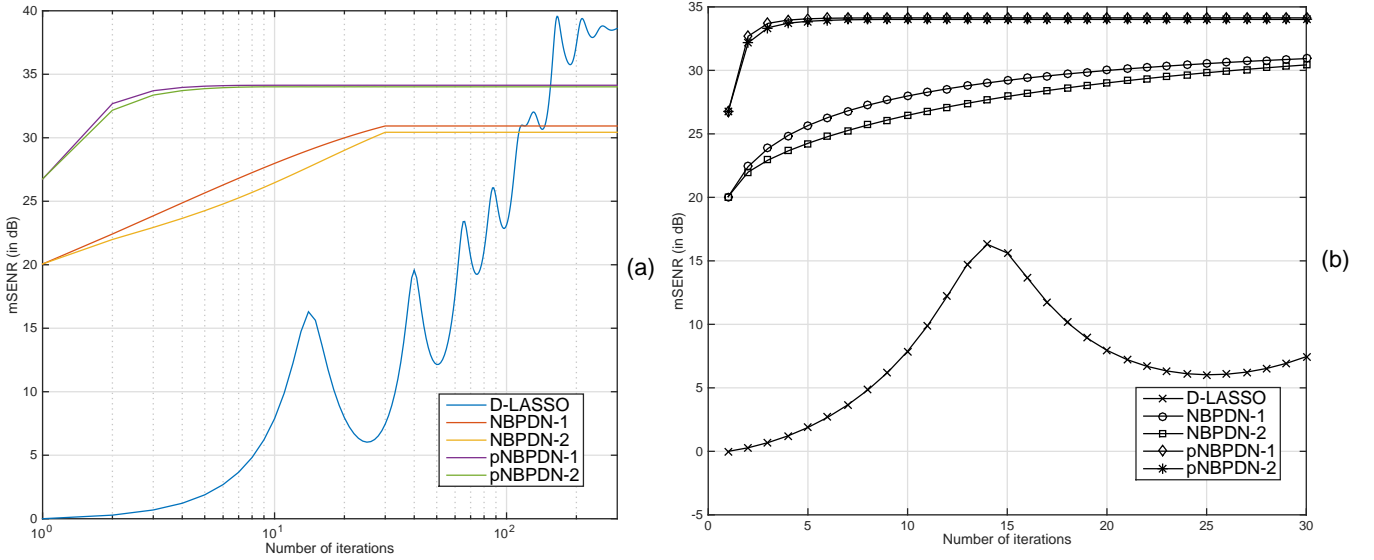}
\caption{Performance of various algorithms with respect to number of iterations over network (number of information exchanging iterations). We set $M = 100$, $N = 500$, $s = 20$, $L = 20$, $d = 4$, and $\text{SNR} = 30$dB. (a) Performances are shown for 300 iterations where we use logarithmic scale to show iterations. (b) Performances are shown for 30 iterations to show quick convergence of proposed algorithms compared to D-LASSO.}
\label{fig:msenr_iter_plot}
\end{figure*}

In this experiment, we observe how fast the algorithms converge with iterations. We set SNR = 30dB and the maximum number of iterations be 300. The results are shown in Fig.~\ref{fig:msenr_iter_plot}. In Fig.~\ref{fig:msenr_iter_plot} (a), we show performance of all our four proposed algorithms and D-LASSO. We assumed that the network can support more communication such that D-LASSO can continue for 300 iterations. We see that D-LASSO has a slow convergence, but provides best performance at the end of iterations. This is expected as it solves the centralized sparse learning problem using ADMM. Among the four proposed algorithms, we see that pNBPDN-1 and 2 perform better, that means a-priori knowledge of sparsity level helps. The pruning based pNBPDN algorithms provide better performance in the sense of quick convergence as well as higher mSENR. In Fig.~\ref{fig:msenr_iter_plot} (b), we show the enlarged portion for initial 30 iterations of Fig.~\ref{fig:msenr_iter_plot} (a). Here it is clear that the proposed algorithms are fast in convergence compared to D-LASSO. 

\subsection{Experiment on Robustness to Measurement Noise}

\begin{figure*}
\hspace{-10mm}
\includegraphics[scale=0.44]{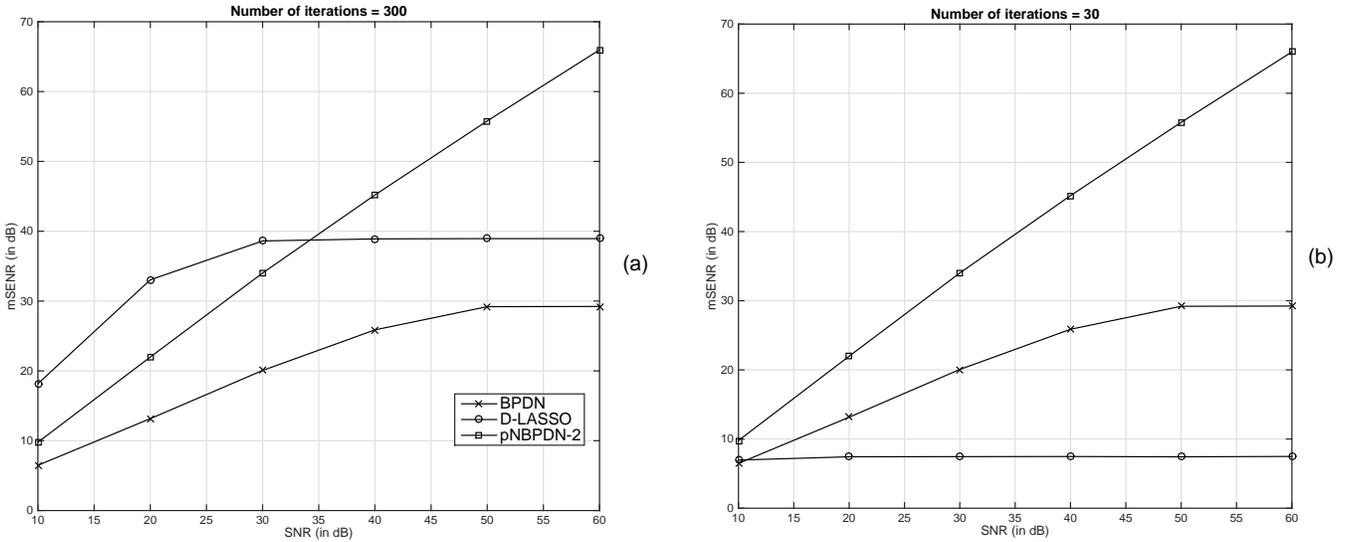}
\caption{Performance comparison of BPDN, D-LASSO and pNBPDN-2 algorithms with respect to SNR. We set M = 100, N = 500, s = 20, L = 20, d = 4. (a) For maximum number of iterations 300. (b) For maximum number of iterations 30.}
\label{fig:msenr_snr_plot}
\end{figure*}

In this experiment, we investigate performance of the algorithms at various SNR conditions to check robustness of algorithms to observation noise power. We show performance of D-LASSO, pNBPDN-2 and BPDN. BPDN does not cooperate over the network. Performances are shown in Fig.~\ref{fig:msenr_snr_plot}. We set allowable numbers of iterations as 300 and 30. Note in the Fig.~\ref{fig:msenr_snr_plot} (a) that D-LASSO hits a floor with increase in SNR. On the other hand pNBDN-2 shows improving performance with the increase in SNR. In the Fig.~\ref{fig:msenr_snr_plot} (b), we allow 30 iterations to simulate a communication constrained case and limited processing time. In this case D-LASSO turns out to be even poorer than BPDN. On the other hand, the pNBDN-2 shows good results for limited iterations. We have simulated other NBPDN algorithms and observed similar trend in performances. Therefore we do not repeat to show those results.

\subsection{Experiment on Sensitivity to Parameter $\lambda$}
\begin{figure}
\hspace{-5mm}
\includegraphics[scale=0.46]{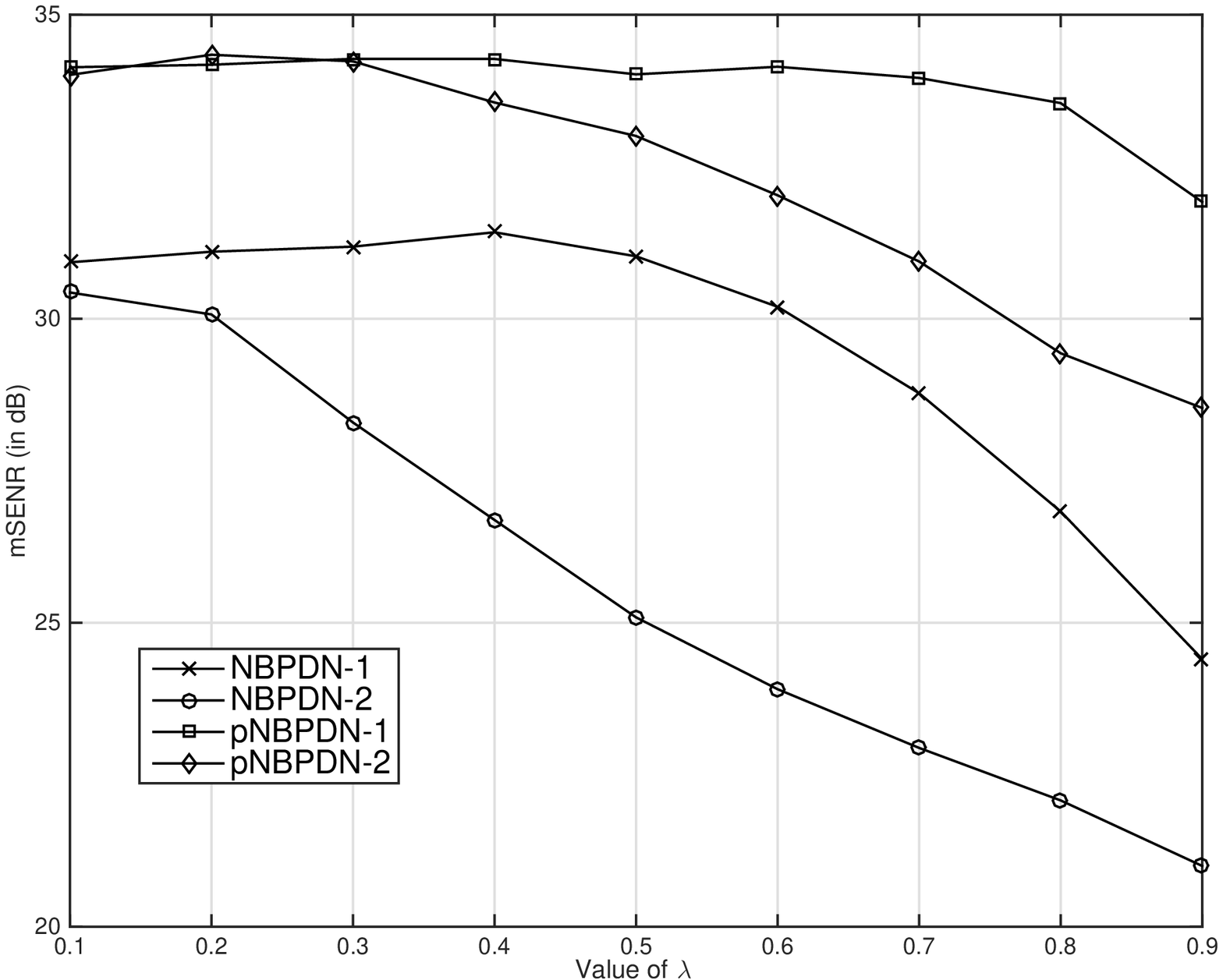}
\caption{Performance of algorithms with respect to the parameter $\lambda$. We set $M = 100$, $N = 500$, $s = 20$, $L = 20$, $d = 4$, and $SNR = 30 \, dB$.}
\label{fig:msenr_alpha_plot}
\end{figure}

In all our previous experiments we have set the parameter $\lambda = 0.1$ for the proposed algorithms. This value was chosen based on an experiment where we vary $\lambda$ for SNR = 30 dB and plot the performance in Fig.~\ref{fig:msenr_alpha_plot}. It can be seen that the chosen $\lambda$ is in the good performance region. 
A higher value of $\lambda$ provides more weight to $\|\mathbf{x}\|_{1}$ and less weight to $g(\mbx,\{\hat{\mbx}_{r,k-1}, h_{lr}\})$. That means, more weight is assigned to sparsity promotion and less weight in using information from neighbors. An important observation is that when $g(.)$ function is $\ell_1$-norm based then the performance is less sensitive to the $\lambda$ change compared to the $\ell_2$-norm case.
Recall that in section~\ref{subsec:NBPDN}, we had a hypothesis as follows: $\ell_1$-norm based $g(.)$ function promotes additional level of sparsity which is not the case for $\ell_2$-norm based $g(.)$ function. It can be seen that the simulation results commensurate with our hypothesis.

\subsection{Reproducible results}
In the spirit of reproducible research, we provide relevant Matlab codes at https://www.kth.se/ise/research/reproducibleresearch-1.433797 and https://sites.google.com/site/saikatchatt/. The code produces the results shown in the figures.

\section{Conclusion}
\label{sec:conclusion}
We show that locally convex algorithms are good for a distributed learning setup where a sparse signal is estimated over a network. The locally convex algorithms are fast in convergence, saving communication and computing resources. It is important to engineer appropriate regularization constraints for the locally convex algorithms. The algorithms are robust with additive noise model, and have theoretical support on their performance. Our theoretical analysis shows that the restricted-isometry-property based estimation guarantees of proposed algorithms and the basis pursuit denoising (BPDN) algorithm are similar.

\section{Theoretical proofs}
\label{sec:Theoretical_proofs}

\subsection{Proof of Lemma~\ref{lem:l1_l2_bound_pruning}}
The $\ell_2$ bounds follow from \cite[Lemma 2]{Liu_improved_SP_analysis_SPL_2014}. We next show the $\ell_1$ bounds. From the inequality between $\ell_1$ and $\ell_2$ norms, we have
\begin{eqnarray*}
\|\left(\mbx-\bar{\mbx}\right)_{\mathcal{S}}\|_1 \leq \sqrt{s_2}\|\left(\mbx-\bar{\mbx}\right)_{\mathcal{S}}\|.
\end{eqnarray*}
The first $\ell_1$ bound follows by using the above inequality in the first $\ell_2$ bound and the fact that $\|.\| \leq \|.\|_1$. Next, as $\|\mbx-\bar{\mbx}\|_0 \leq s_1+s_2$, we can write
\begin{eqnarray*}
\|\mbx-\bar{\mbx}\|_1 \leq \sqrt{s_1+s_2}\|\mbx-\bar{\mbx}\|.
\end{eqnarray*}
The second $\ell_1$ bound follows directly by substituting the above bound in the second $\ell_2$ bound. \QEDB

\subsection{Proof of Lemma~\ref{lem:smaller_indices_bound_l1_l2}}
The $\ell_2$ bounds follow from [improved SP, lemma 3]. We next show the $\ell_1$ bound.
Consider the following relation,
\begin{eqnarray*}
\begin{array}{rl}
\|{\mathbf{z}}_{\mathcal{S}_1\cap\mathcal{S}_{\nabla}}\|_1 & =  \|\mbx_{\mathcal{S}_1\cap\mathcal{S}_{\nabla}} + \left({\mathbf{z}}-\mbx\right)_{\mathcal{S}_1\cap\mathcal{S}_{\nabla}}\|_1 \\
& \geq \|\mbx_{\mathcal{S}_1\cap\mathcal{S}_{\nabla}}\|_1 -\| \left({\mathbf{z}}-\mbx\right)_{\mathcal{S}_1\cap\mathcal{S}_{\nabla}}\|_1.
\end{array}
\end{eqnarray*}
Rearranging the terms in the above equation gives
\begin{eqnarray}
\label{eq:index_lemma_bound_l2_1}
\begin{array}{rl}
\|\mbx_{\mathcal{S}_1\cap\mathcal{S}_{\nabla}}\|_1 & \leq \|{\mathbf{z}}_{\mathcal{S}_1\cap\mathcal{S}_{\nabla}}\|_1 + \| \left({\mathbf{z}}-\mbx\right)_{\mathcal{S}_1\cap\mathcal{S}_{\nabla}}\|_1.
\end{array}
\end{eqnarray}
Define $\acute{\mathcal{S}} \triangleq \mathcal{S}_2 \setminus \mathcal{S}_{\nabla}$ as the set of indices of the $s_1$ highest magnitude elements in ${\mathbf{z}}$. Now, we can write,
\begin{eqnarray*}
\begin{array}{rl}
\|{\mathbf{z}}_{\mathcal{S}_1\cap\mathcal{S}_{\nabla}}\|_1 & = \|{\mathbf{z}}_{\mathcal{S}_1\cap\mathcal{S}_{\nabla}}\|_1 + \|{\mathbf{z}}_{\acute{\mathcal{S}}}\|_1 - \|{\mathbf{z}}_{\acute{\mathcal{S}}}\|_1 \\
& \leq \|{\mathbf{z}}_{\mathcal{S}_2}\|_1 -\|{\mathbf{z}}_{\acute{\mathcal{S}}}\|_1 \\
& = \|{\mathbf{z}}_{\mathcal{S}_2\setminus\mathcal{S}_1}\|_1 + \|{\mathbf{z}}_{\mathcal{S}_1}\|_1 - \|{\mathbf{z}}_{\acute{\mathcal{S}}}\|_1 \\
& \leq \|{\mathbf{z}}_{\mathcal{S}_2\setminus\mathcal{S}_1}\|_1,
\end{array}
\end{eqnarray*}
where we used the highest magnitude property of $\acute{\mathcal{S}}$ in the last step. The above equation can be written as
\begin{eqnarray}
\label{eq:index_lemma_bound_l2_2}
\begin{array}{l}
\|{\mathbf{z}}_{\mathcal{S}_1\cap\mathcal{S}_{\nabla}}\|_1  \leq \|{\mathbf{z}}_{\mathcal{S}_2\setminus \mathcal{S}_1}\|_1 = \|\left( {\mathbf{z}}- \mbx \right)_{\mathcal{S}_2\setminus\mathcal{S}_1}\|_1.
\end{array}
\end{eqnarray}
From \eqref{eq:index_lemma_bound_l2_1} and \eqref{eq:index_lemma_bound_l2_2}, and from the fact that $(\mathcal{S}_2\setminus\mathcal{S}_1)\cap(\mathcal{S}_1\cap\mathcal{S}_{\nabla}) = \emptyset$, we have
\begin{eqnarray*}
\begin{array}{rl}
\|\mbx_{\mathcal{S}_{\nabla}}\|_1 & = \|\mbx_{\mathcal{S}_1\cap\mathcal{S}_{\nabla}}\|_1 \\
& \leq \|\left(  \mbx - {\mathbf{z}} \right)_{\mathcal{S}_2\setminus\mathcal{S}_1}\|_1 + \| \left(\mbx - {\mathbf{z}}\right)_{\mathcal{S}_1\cap\mathcal{S}_{\nabla}}\|_1 \\
& \leq \|\left(\mbx - {\mathbf{z}}\right)_{\mathcal{S}_2}\|_1.
\end{array}
\end{eqnarray*} \QEDB

\subsection{Proof of Lemma~\ref{lem:inequality_lemma}}
Denote $\acute{\mbx}_{l,k} = \sum\limits_{r \in \mathcal{N}_l} h_{lr} \hat{\mbx}_{r,k-1}$. The solution, $\hat{\mbx}_{l,k} \left(= \mbx + \mbz_{l,k}\right)$ follows the inequality,
\begin{eqnarray*}
\begin{array}{rl}
&\hspace{-4mm} \lambda \|\mbx\|_1 + (1-\lambda) \|\mbx-\acute{\mbx}_{l,k}\|_1 \\
&\geq \lambda \|\mbx + \mbz_{l,k}\|_1 + (1-\lambda) \|\mbx + \mbz_{l,k} - \acute{\mbx}_{l,k}\|_1 \\
&\stackrel{(a)}{\geq} \lambda \|(\mbx + \mbz_{l,k})_{\T_0}\|_1 + \lambda \|(\mbx + \mbz_{l,k})_{\T_0^c}\|_1 \\ 
& \hspace{5mm} - (1-\lambda) \|\mbz_{l,k} \|_1 + (1-\lambda) \|\mbx  - \acute{\mbx}_{l,k}\|_1 \\
& \hspace{-5mm} \stackrel{(b)}{\geq} \lambda\|\mbx_{\T_0}\|_1 - \lambda\|\left(\mbz_{l,k}\right)_{\T_0}\|_1 + \lambda\|\left(\mbz_{l,k}\right)_{\T_0^c}\|_1 -                 \lambda\|\mbx_{\T_0^c}\|_1 \\
& \hspace{-5mm} - (1-\lambda) \left[\|\left(\mbz_{l,k}\right)_{\T_0^c} \|_1+ \|\left(\mbz_{l,k}\right)_{\T_0} \|_1\right] + (1-\lambda) \|\mbx  - \acute{\mbx}_{l,k}\|_1 ,
\end{array}
\end{eqnarray*}
where $(a)$ and $(b)$ follows from the fact that $\T_0$, $\T_0^c$ are disjoint and the reverse triangle inequality of the $\ell_1$ norm. It can be seen that \eqref{eq:lemma_algo1_1} follows from rearranging the above inequality.
An alternate inequality reduction can be pursued as follows,
\begin{eqnarray*}
\begin{array}{rl}
&\hspace{-4mm} \lambda \|\mbx\|_1 + (1-\lambda) \|\mbx-\acute{\mbx}_{l,k}\|_1 \\
&\geq \lambda \|\mbx + \mbz_{l,k}\|_1 + (1-\lambda) \|\mbx + \mbz_{l,k} - \acute{\mbx}_{l,k}\|_1 \\
&\geq \lambda \|(\mbx + \mbz_{l,k})_{\T_0}\|_1 + \lambda \|(\mbx + \mbz_{l,k})_{\T_0^c}\|_1 \\ 
& \hspace{5mm} + (1-\lambda) \|\mbz_{l,k} \|_1 - (1-\lambda) \|\mbx  - \acute{\mbx}_{l,k}\|_1 \\
& \hspace{-5mm} \geq \lambda\|\mbx_{\T_0}\|_1 - \lambda\|\left(\mbz_{l,k}\right)_{\T_0}\|_1 + \lambda\|\left(\mbz_{l,k}\right)_{\T_0^c}\|_1 -                 \lambda\|\mbx_{\T_0^c}\|_1 \\
& \hspace{-5mm} + (1-\lambda) \left[\|\left(\mbz_{l,k}\right)_{\T_0^c} \|_1+ \|\left(\mbz_{l,k}\right)_{\T_0} \|_1\right] - (1-\lambda) \|\mbx  - \acute{\mbx}_{l,k}\|_1.
\end{array}
\end{eqnarray*}
Now, \eqref{eq:lemma_algo1_2} follows from the above inequality. \QEDB

\subsection{Proof of Theorem~\ref{thm:l1_conv_theorem1}}
The proof follows similar structure as adopted in \cite[Theorem 2]{Cai_Shifting_Inequality_TSP_2010}. For notational simplicity we drop the subscripts $\{l,k\}$. We can write
\begin{eqnarray}
\label{eq:l1_conv_inequality1}
\begin{array}{rl}
\sum\limits_{i \geq 1} \|\mbz_{\T_i}\| & \hspace{-3mm}\stackrel{(a)}{\leq}  \frac{\|\mbz_{\T_0^c}\|_1}{\sqrt{b}} \stackrel{(b)}{\leq} \frac{\|\mbz_{\T_0}\|_1}{\sqrt{b}(2\lambda -1)} + \frac{2\lambda}{\sqrt{b}(2\lambda -1)}\|\mbx_{\T_0^c}\|_1 \\
& \hspace{-3mm} \stackrel{(c)}{\leq} \frac{\sqrt{k}}{\sqrt{b}(2\lambda -1)}\|\mbz_{\T_0 \cup \T_*}\| + \frac{2\lambda}{\sqrt{b}(2\lambda -1)}\|\mbx_{\T_0^c}\|_1,
\end{array}
\end{eqnarray}
where $(a)$ and $(b)$ follow from Lemma~\ref{lem:shifting_inequality_lemma} and \eqref{eq:lemma_algo1_1}; and $(c)$ follows from the bounding of $\ell_1$-norm by the $\ell_2$-norm. Next, consider
\begin{eqnarray}
\label{eq:bound_equation_4}
\begin{array}{l}
\|\mbA \mbz_{\T_0 \cup \T_*}\|^2 \\
 =  \left\langle \mbA\mbz_{\T_0 \cup \T_*},\mbA \mbz\right\rangle - \langle \mbA \mbz_{\T_0 \cup \T_*},\sum_{j \geq 1}\mbA \mbz_{\T_j}\rangle \\
 \leq  |\left\langle \mbA\mbz_{\T_0 \cup \T_*},\mbA \mbz\right\rangle| + |\langle \mbA\mbz_{\T_0 \cup \T_*},\sum_{j \geq 1}\mbA \mbz_{\T_j}\rangle|.
 \end{array}
\end{eqnarray}
The first term in the above equation can be written as
\begin{eqnarray*}
\begin{array}{rl}
\hspace{-3mm}|\left\langle \mbA\mbz_{\T_0 \cup \T_*},\mbA \mbz\right\rangle|  
& \leq  \|\mbA\mbz_{\T_0 \cup \T_*}\| \|\mbA \mbz\|\\
&  \stackrel{(a)}{\leq} 2 \epsilon \sqrt{1+\delta_{s+a}} \|\mbz_{\T_0 \cup \T_*}\|,
\end{array}
\end{eqnarray*}
where $(a)$ follows from the RIP of $\mbA$ and \eqref{eq:h_bound_eq_1}. Also, the second term in \eqref{eq:bound_equation_4} can be bounded as
\begin{eqnarray*}
\begin{array}{l}
\langle \mbA\mbz_{\T_0 \cup \T_*},\sum_{j \geq 1}\mbA \mbz_{\T_j}\rangle 
\stackrel{(a)}{\leq} \theta_{s+a,b} \|\mbz_{\T_0 \cup \T_*}\| \sum_{j\geq 1} \|\mbz_{\T_j}\| \\
\hspace{5mm} \stackrel{(b)}{\leq} \sqrt{\frac{s}{b}}\frac{\theta_{s+a,b}}{2\lambda -1}\|\mbz_{\T_0 \cup \T_*}\|^2 + \frac{2\lambda \theta_{s+a,b}}{\sqrt{b}(2\lambda -1)}\|\mbz_{\T_0 \cup \T_*}\| \|\mbx_{\T_0^c}\|_1,
\end{array}
\end{eqnarray*}
where $(a)$ follows from Definition~\ref{def:rop} and $(b)$ follows from \eqref{eq:l1_conv_inequality1}. Using the RIP and the above two inequalities, we can write,
\begin{eqnarray}
\label{eq:l1_conv_inequality2}
\begin{array}{rl}
(1-\delta_{s+a})\|\mbz_{\T_0 \cup \T_*}\|^2 & \hspace{-3mm} \leq \|\mbA\mbz_{\T_0 \cup \T_*}\|^2 \\ & \hspace{-30mm} \leq \|\mbz_{\T_0 \cup \T_*}\| (2\epsilon \sqrt{1+\delta_{s+a}} + \theta_{s+a,b} \sum_{j\geq 1} \|\mbz_{\T_j}\| ).
\end{array}
\end{eqnarray}
Using \eqref{eq:l1_conv_inequality1}, the above equation can be simplified as
\begin{eqnarray}
\label{eq:h_bound_eq_3}
\begin{array}{l}
\|\mbz_{\T_0 \cup \T_*}\| \leq \frac{2\sqrt{1+\delta_{s+a}}}{c_1} \epsilon + \frac{2 \lambda \theta_{s+a,b}}{(2\lambda-1)c_1 \sqrt{b}} \|\mbx_{\T_0^c}\|_1.
\end{array}
\end{eqnarray}
Now, we can upper bound the error as
\begin{eqnarray*}
\begin{array}{l}
\|\mbz\| \leq \|\mbz\|_1 = \|\mbz_{\T_0}\|_1 + \|\mbz_{\T_0^c}\|_1 \\
\hspace{3mm} \stackrel{(a)}{\leq} \frac{2\lambda \sqrt{s}}{2\lambda-1}\|\mbz_{\T_0 \cup \T_*}\| + \frac{2\lambda}{2\lambda-1}\|\mbx_{\T_0^c}\|_1 \\
\hspace{3mm} \stackrel{(b)}{\leq} \frac{4 \lambda \sqrt{s(1+\delta_{s+a})}}{c_1(2\lambda -1)} \epsilon + \frac{2\lambda}{2\lambda -1}\left[1 + \frac{2\lambda}{2\lambda -1} \sqrt{\frac{s}{b}} \frac{\theta_{s+a,b}}{c_1}\right]\|\mbx_{\T_0^c}\|_1,
\end{array}
\end{eqnarray*}
where $(a)$ follows from \eqref{eq:lemma_algo1_1} and bounding $\ell_1$-norm by the $\ell_2$-norm. Also, $(b)$ follows from \eqref{eq:h_bound_eq_3}. The above bound on the estimation error is valid at every iteration $k$ as long as $c_1 > 0$ which reduces to the condition $\delta_{s+a}+\sqrt{\frac{s}{b}}\frac{\theta_{s+a,b}}{2\lambda -1} < 1$.
\QEDB

\subsection{Proof of Theorem~\ref{thm:l1_recurrence_theorem2}}
The first part of the proof follows as in Theorem~\ref{thm:l1_conv_theorem1}. We can write
\begin{eqnarray*}
\label{eq:l1_recurrence_inequality1}
\begin{array}{rl}
\sum\limits_{i \geq 1} \|\mbz_{\T_i}\| & \stackrel{(a)}{\leq}  \frac{\|\mbz_{\T_0^c}\|_1}{\sqrt{b}} \\ 
& \hspace{-11mm} \stackrel{(b)}{\leq} \frac{2 \lambda -1}{\sqrt{b}}\|\mbz_{\T_0}\|_1 + \frac{2\lambda}{\sqrt{b}}\|\mbx_{\T_0^c}\|_1 + \frac{2(1-\lambda)}{\sqrt{b}}\|\mbx  - \acute{\mbx}_{l,k}\|_1 \\
& \hspace{-11mm} \stackrel{(c)}{\leq} \frac{(2 \lambda' -1)\sqrt{k}}{\sqrt{b}}\|\mbz_{\T_0 \cup \T_*}\| + \frac{2\lambda}{\sqrt{b}}\|\mbx_{\T_0^c}\|_1 + \frac{2(1-\lambda)}{\sqrt{b}}\|\mbx  - \acute{\mbx}_{l,k}\|_1,
\end{array}
\end{eqnarray*}
where $(a)$ and $(b)$ follow from Lemma~\ref{lem:shifting_inequality_lemma} and \eqref{eq:lemma_algo1_2}; and $(c)$ follows from the bounding of $\ell_1$-norm by the $\ell_2$-norm and $\lambda' = \max\{\lambda,1/2\}$. Plugging the above inequality in  \eqref{eq:l1_conv_inequality2}, we have
\begin{eqnarray}
\label{eq:l1_recurrence_inequality2}
\begin{array}{rl}
\|\mbz_{\T_0 \cup \T_*}\|  & \hspace{-3mm}\leq  \frac{2\sqrt{1+\delta_{s+a}}}{c_5} \epsilon \\ 
& \hspace{-2mm}+ \frac{2 \lambda \theta_{s+a,b}}{c_5 \sqrt{b}} \|\mbx_{\T_0^c}\|_1 + \frac{2(1-\lambda)\theta_{s+a,b}}{c_5 \sqrt{b}}\|\mbx  - \acute{\mbx}_{l,k}\|_1,
\end{array}
\end{eqnarray}
where $c_5 = 1-\delta_{s+a}-\sqrt{\frac{s}{b}} (2\lambda' -1)\theta_{s+a,b} > 0$. The error can now be bounded as
\begin{eqnarray*}
\begin{array}{rl}
\|\mbz\|_1 & = \|\mbz_{\T_0}\|_1 + \|\mbz_{\T_0^c}\|_1 \\
& \stackrel{(a)}{\leq} 2\lambda \sqrt{s}\|\mbz_{\T_0 \cup \T_*}\| + 2\lambda\|\mbx_{\T_0^c}\|_1 + 2(1-\lambda)\|\mbx  - \acute{\mbx}_{l,k}\|_1 \\
& \stackrel{(b)}{\leq} c_2 \|\mbx  - \acute{\mbx}_{l,k}\|_1 + c_3 \|\mbx_{\T_0^c}\|_1 + c_4 \epsilon,
\end{array}
\end{eqnarray*}
where $(a)$ follows from Lemma~\ref{lem:inequality_lemma} and $(b)$ follows from \eqref{eq:l1_recurrence_inequality2}. The first term in RHS of the above inequality can be bounded as $\|\mbx  - \acute{\mbx}_{l,k}\|_1 = \|\mbx  - \sum\limits_{r \in \mathcal{N}_l} h_{lr} \hat{\mbx}_{r,k-1}\|_1 \stackrel{(a)}{\leq}  \sum\limits_{r \in \mathcal{N}_l} h_{lr} \|\mbx  - \hat{\mbx}_{r,k-1}\|_1$, where $(a)$ follows from the right stochastic property of $\mbH$. It can be seen that the result follows by substitution. \QEDB

\subsection{Proof of Theorem~\ref{thm:l1_convergence_theorem2}}
From the condition, we have, $\delta_{s+a}+\sqrt{\frac{s}{b}} (2\lambda'  -1)\theta_{s+a,b} < \delta_{s+a}+\sqrt{\frac{s}{b}}\theta_{s+a,b} < 1$ for $\lambda' < 1$. Hence, from Theorem~\ref{thm:l1_recurrence_theorem2} we can write
\begin{eqnarray*}
\|\mbz_{l,k}\|_1\leq c_2 \sum\limits_{r \in \mathcal{N}_l} h_{lr} \|\mbz_{r,k-1}\|_1+ c_3 \|\mbx_{\T_0^c}\|_1 + c_4 \epsilon.
\end{eqnarray*}
Let $\underline{\|\mbz_{k}\|_1} \triangleq \left[ \|\mbz_{1,k}\|_1 \hdots \|\mbz_{L,k}\|_1\right]^{t}$. We can vectorize the above equation to write
\begin{eqnarray*}
\underline{\|\mbz_{k}\|_1}\leq c_2 \mbH \underline{\|\mbz_{k-1}\|_1} + c_3 \|\mbx_{\T_0^c}\|_1 \mathbf{1}_L + c_4 \epsilon \mathbf{1}_L.
\end{eqnarray*}
Applying the above equation iteratively, we have
\begin{eqnarray}
\label{eq:l1_conv_thm_eq1}
\begin{array}{rl}
\underline{\|\mbz_{k}\|_1}& \leq c_2^k \mbH^k \underline{\|\mbz_{0}\|_1} \\ 
& \hspace{0mm}+ \left[ \mathbf{I}_L + \hdots + (c_2 \mbH)^{k-1}\right](c_3 \|\mbx_{\T_0^c}\|_1 + c_4 \epsilon )\mathbf{1}_L,
\end{array}
\end{eqnarray}
where $\underline{\|\mbz_{0}\|_1} \triangleq \left[ \|\mbx-\hat{\mbx}_{1,0}\|_1 \hdots \|\mbx-\hat{\mbx}_{L,0}\|_1\right]^{t}$ is the estimation error in the intialization step. Note that BPDN algorithm is used in the initialization step which is the modified algorithm with $\lambda = 1$. From Theorem~\ref{thm:l1_recurrence_theorem2}, we have at node $l$ (and $\lambda = 1$)
\begin{eqnarray*}
\|\mbx-\hat{\mbx}_{l,0}\|_1 \leq c_6 \|\mbx_{\T_0^c}\| + c_7 \epsilon,
\end{eqnarray*}
where $c_6 = 2 \left[1 + 2 \sqrt{\frac{s}{b}} \frac{\theta_{s+a,b}}{c_5}\right], c_7 = \frac{4 \sqrt{s(1+\delta_{s+a})}}{c_5}$. Now substituting the above inequality in \eqref{eq:l1_conv_thm_eq1} and using the power series reduction, we get
\begin{eqnarray*}
\underline{\|\mbz_{k}\|_1} \leq \frac{c_2^k-1}{c_2-1}\left[(c_3+c_2c_6)\|\mbx_{\T_0^c}\| + (c_4+c_2c_7)\epsilon\right]
\end{eqnarray*}
which gives the result. Observe that we have used the right stochastic property of $\mbH$ to reduce the power series in the above inequality. The condition for the above bound to hold is that $\delta_{s+a}+\sqrt{\frac{s}{b}}\theta_{s+a,b} < 1$. \QEDB

\subsection{Proof of Lemma~\ref{lem:inequality_lemma_algo5}}
The first part of the proof is similar to Lemma~\ref{lem:inequality_lemma}. Denote $\acute{\mbx}_{l,k} = \sum\limits_{r \in \mathcal{N}_l} h_{lr} \hat{\mbx}_{r,k-1}$. The solution, $\hat{\mbx}_{l,k} \left(= \mbx + \mbz_{l,k}\right)$ follows the inequality,
\begin{eqnarray*}
\begin{array}{rl}
&\hspace{-4mm} \lambda \|\mbx\|_1 + (1-\lambda) \|\mbx-\acute{\mbx}_{l,k}\| \\
&\geq \lambda \|\mbx + \mbz_{l,k}\|_1 + (1-\lambda) \|\mbx + \mbz_{l,k} - \acute{\mbx}_{l,k}\| \\
&\stackrel{(a)}{\geq} \lambda \|(\mbx + \mbz_{l,k})_{\T_0}\|_1 + \lambda \|(\mbx + \mbz_{l,k})_{\T_0^c}\|_1 \\ 
& \hspace{5mm} - (1-\lambda) \|\mbz_{l,k} \| + (1-\lambda) \|\mbx  - \acute{\mbx}_{l,k}\| \\
& \hspace{-5mm} \stackrel{(b)}{\geq} \lambda\|\mbx_{\T_0}\|_1 - \lambda\|\left(\mbz_{l,k}\right)_{\T_0}\|_1 + \lambda\|\left(\mbz_{l,k}\right)_{\T_0^c}\|_1 -                 \lambda\|\mbx_{\T_0^c}\|_1 \\
& \hspace{-5mm} - (1-\lambda) \left[\|\left(\mbz_{l,k}\right)_{\T_0^c} \|_1+ \|\left(\mbz_{l,k}\right)_{\T_0} \|_1\right] + (1-\lambda) \|\mbx  - \acute{\mbx}_{l,k}\| ,
\end{array}
\end{eqnarray*}
where $(a)$ and $(b)$ follows from the fact that $\T_0$, $\T_0^c$ are disjoint and the reverse triangle inequality of the $\ell_1$ norm. It can be seen that \eqref{eq:lemma_algo1_1} follows from rearranging the above inequality.
The second inequality can be derived as follows,
\begin{eqnarray*}
\begin{array}{rl}
&\hspace{-4mm} \lambda \|\mbx\|_1 + (1-\lambda) \|\mbx-\acute{\mbx}_{l,k}\| \\
&\geq \lambda \|\mbx + \mbz_{l,k}\|_1 + (1-\lambda) \|\mbx + \mbz_{l,k} - \acute{\mbx}_{l,k}\| \\
&\geq \lambda \|(\mbx + \mbz_{l,k})_{\T_0}\|_1 + \lambda \|(\mbx + \mbz_{l,k})_{\T_0^c}\|_1 \\ 
& \hspace{5mm} + (1-\lambda) \|\mbz_{l,k} \| - (1-\lambda) \|\mbx  - \acute{\mbx}_{l,k}\| \\
& \hspace{-5mm} \geq \lambda\|\mbx_{\T_0}\|_1 - \lambda\|\left(\mbz_{l,k}\right)_{\T_0}\|_1 + \lambda\|\left(\mbz_{l,k}\right)_{\T_0^c}\|_1 -                 \lambda\|\mbx_{\T_0^c}\|_1 \\
& \hspace{-5mm} + (1-\lambda) \|\left(\mbz_{l,k}\right)_{\T_0} \| - (1-\lambda) \|\mbx  - \acute{\mbx}_{l,k}\|.
\end{array}
\end{eqnarray*}
Now, \eqref{eq:lemma_algo1_2} follows from the above inequality by using the relation $\|\left(\mbz_{l,k}\right)_{\T_0} \| \geq \frac{1}{\sqrt{s}}\|\left(\mbz_{l,k}\right)_{\T_0} \|_1$.
\QEDB

\subsection{Proof of Theorem~\ref{thm:l2_conv_theorem1}}
The first part follows from the proof of Theorem~\ref{thm:l1_conv_theorem1}. It can be seen that for $\lambda > 1/2$, \eqref{eq:lemma_algo5_1} is valid. Hence, we can write from \eqref{eq:h_bound_eq_3},
\begin{eqnarray}
\label{eq:l2_h_bound_eq_1}
\begin{array}{l}
\|\mbz_{\T_0 \cup \T_*}\| \leq \frac{2\sqrt{1+\delta_{s+a}}}{c_1} \epsilon + \frac{2 \lambda \theta_{s+a,b}}{(2\lambda-1)c_1 \sqrt{b}} \|\mbx_{\T_0^c}\|_1,
\end{array}
\end{eqnarray}
where $c_1 = 1-\delta_{s+a}-\sqrt{\frac{s}{b}}\frac{\theta_{s+a,b}}{2\lambda -1} > 0$. Now, we can upper bound the error as
\begin{eqnarray*}
\begin{array}{l}
\|\mbz\| \leq \|\mbz_{\T_0}\| + \|\mbz_{\T_0^c}\|  \stackrel{\eqref{eq:lemma_algo5_1}}{\leq} \frac{2\lambda\sqrt{s}}{2\lambda -1} \|\mbz_{\T_0}\| + \frac{2\lambda}{2\lambda -1} \|\mbx_{\T_0^c}\|_1 \\
\hspace{5mm} \stackrel{(a)}{\leq} \frac{4 \lambda \sqrt{s(1+\delta_{s+a})}}{c_1(2\lambda -1)} \epsilon + \frac{2\lambda}{2\lambda -1}\left[1 + \frac{2\lambda}{2\lambda -1}  \frac{\sqrt{s}\theta_{s+a,b}}{c_1 \sqrt{b}}\right]\|\mbx_{\T_0^c}\|_1,
\end{array}
\end{eqnarray*}
where $(a)$ is due to the inequality $\|\mbz_{\T_0}\| \leq \|\mbz_{\T_0 \cup \T_*}\|$ and \eqref{eq:l2_h_bound_eq_1}. The above bound on the estimation error is valid at every iteration $k$ as long as $c_1 > 0$ which reduces to the condition $\delta_{s+a}+\sqrt{\frac{s}{b}}\frac{\theta_{s+a,b}}{2\lambda -1} < 1$. \QEDB

\subsection{Proof of Theorem~\ref{thm:l2_recurrence_theorem2}}
From first part of Theorem~\ref{thm:l1_recurrence_theorem2} and \eqref{eq:lemma_algo5_2}, we can write
\begin{eqnarray}
\label{eq:l2_recurrence_inequality1}
\begin{array}{l}
\hspace{-2mm}\sum\limits_{i \geq 1} \|\mbz_{\T_i}\| \\ \hspace{-2mm} \leq \frac{ \lambda(1+\sqrt{s}) -1}{\lambda\sqrt{sb}}\|\mbz_{\T_0}\|_1 + \frac{2}{\sqrt{b}}\|\mbx_{\T_0^c}\|_1 + \frac{2(1-\lambda)}{\lambda \sqrt{b}}\|\mbx  - \acute{\mbx}_{l,k}\| \\
\hspace{-2mm} \stackrel{(a)}{\leq} \frac{\lambda^{\prime \prime}(1+\sqrt{s}) -1}{\sqrt{b}}\|\mbz_{\T_0 \cup \T_*}\| + \frac{2}{\sqrt{b}}\|\mbx_{\T_0^c}\|_1 + \frac{2(1-\lambda)}{\lambda\sqrt{b}}\|\mbx  - \acute{\mbx}_{l,k}\|,
\end{array}
\end{eqnarray}
where $(a)$ follows from the bounding of $\ell_1$-norm by the $\ell_2$-norm and $\lambda' = \max\left\{\lambda,\frac{1}{1+\sqrt{s}}\right\}$. Plugging the above inequality in  \eqref{eq:l1_conv_inequality2}, we have
\begin{eqnarray}
\label{eq:l2_recurrence_inequality2}
\begin{array}{rl}
\|\mbz_{\T_0 \cup \T_*}\| & \hspace{-2mm}\leq \frac{2\sqrt{1+\delta_{s+a}}}{c_8} \epsilon \\
& + \frac{2\theta_{s+a,b}}{c_8 \sqrt{b}} \|\mbx_{\T_0^c}\|_1 + \frac{2(1-\lambda)\theta_{s+a,b}}{c_8 \lambda \sqrt{b}}\|\mbx  - \acute{\mbx}_{l,k}\|,
\end{array}
\end{eqnarray}
where $c_8 = 1-\delta_{s+a}-\frac{\lambda^{\prime \prime}(1+\sqrt{s})-1}{\sqrt{b}}\theta_{s+a,b} > 0$. The error can now be bounded as
\begin{eqnarray*}
\begin{array}{l}
\|\mbz\|  \leq \|\mbz_{\T_0 \cup \T_*}\| + \sum\limits_{i \geq 1} \|\mbz_{\T_i}\| \\
 \stackrel{(a)}{\leq} \left[1+\frac{\lambda^{\prime \prime}(1+\sqrt{s})-1}{\sqrt{b}}\right]\|\mbz_{\T_0 \cup \T_*}\| + \frac{2}{\sqrt{b}}\|\mbx_{\T_0^c}\|_1\\
  \hspace{5mm} + \frac{2(1-\lambda)}{\lambda \sqrt{b}}\|\mbx  - \acute{\mbx}_{l,k}\| 
 \stackrel{(b)}{\leq} c_9 \|\mbx  - \acute{\mbx}_{l,k}\| + c_{10} \|\mbx_{\T_0^c}\|_1 + c_{11} \epsilon,
\end{array}
\end{eqnarray*}
where $(a)$ follows from \eqref{eq:l2_recurrence_inequality1} and $(b)$ follows from substituting \eqref{eq:l2_recurrence_inequality2}. We can bound the first term in RHS of the above inequality as $\|\mbx  - \acute{\mbx}_{l,k}\| = \|\mbx  - \sum\limits_{r \in \mathcal{N}_l} h_{lr} \hat{\mbx}_{r,k-1}\| \stackrel{(a)}{\leq}  \sum\limits_{r \in \mathcal{N}_l} h_{lr} \|\mbx  - \hat{\mbx}_{r,k-1}\|$, where $(a)$ follows from the right stochastic property of $\mbH$. It can be seen that the recurrence inequality follows by substitution. \QEDB

\subsection{Proof of Theorem~\ref{thm:l2_convergence_theorem2}}
From the condition, we have, $\delta_{s+a}+\sqrt{\frac{s}{b}} (2\lambda^{\prime \prime}  -1)\theta_{s+a,b} < \delta_{s+a}+\sqrt{\frac{s}{b}}\theta_{s+a,b} < 1$ for $\lambda^{\prime \prime} < 1$. Hence, from Theorem~\ref{thm:l2_recurrence_theorem2} we can write
\begin{eqnarray*}
\|\mbz_{l,k}\|\leq c_9 \sum\limits_{r \in \mathcal{N}_l} h_{lr} \|\mbz_{r,k-1}\|+ c_{10} \|\mbx_{T_0^c}\|_1 + c_{11} \epsilon.
\end{eqnarray*}
Let $\underline{\|\mbz_{k}\|} \triangleq \left[ \|\mbz_{1,k}\| \hdots \|\mbz_{L,k}\|\right]^{t}$. With vectorization the above equation can be compactly written as
\begin{eqnarray*}
\underline{\|\mbz_{k}\|}\leq c_9 \mbH \underline{\|\mbz_{k-1}\|} + c_{10} \|\mbx_{T_0^c}\|_1 \mathbf{1}_L + c_{11} \epsilon \mathbf{1}_L.
\end{eqnarray*}
Applying the above equation iteratively, we get
\begin{eqnarray}
\label{eq:l2_conv_thm_eq1}
\begin{array}{rl}
\underline{\|\mbz_{k}\|}& \leq c_9^k \mbH^k \underline{\|\mbz_{0}\|} \\ 
& \hspace{-2mm}+ \left[ \mathbf{I}_L + \hdots + (c_9 \mbH)^{k-1}\right](c_{10} \|\mbx_{T_0^c}\|_1 + c_{11} \epsilon )\mathbf{1}_L,
\end{array}
\end{eqnarray}
where $\underline{\|\mbz_{0}\|} \triangleq \left[ \|\mbx-\hat{\mbx}_{1,0}\| \hdots \|\mbx-\hat{\mbx}_{L,0}\|\right]^{t}$ is the estimation error in the intialization step. Using Lemma~\ref{lem:BPDN_convergence_result} at node $l$, the following bound holds
\begin{eqnarray*}
\|\mbx-\hat{\mbx}_{l,0}\| \leq c_{12} \|\mbx_{\T_0^c}\|_1 + c_{13} \epsilon.
\end{eqnarray*}
Now substituting the above inequality in \eqref{eq:l2_conv_thm_eq1} and using the power series reduction, we get
\begin{eqnarray*}
\underline{\|\mbz_{k}\|} \leq \frac{c_9^k-1}{c_9-1}\left[(c_{10}+c_9 c_{12})\|\mbx_{\T_0^c}\|_1 + (c_{11}+c_9 c_{13})\epsilon\right]
\end{eqnarray*}
which gives the result. Note that we have used the right stochastic property of $\mbH$ to reduce the power series in the above inequality. The condition for the above bound to hold is that $\delta_{s+a}+\sqrt{\frac{s}{b}}\theta_{s+a,b} < 1$.  \QEDB

\subsection{Proof of Lemma~\ref{lem:pruning_bound_lemma1}}
We will use Lemma~\ref{lem:l1_l2_bound_pruning} and Lemma~\ref{lem:smaller_indices_bound_l1_l2} to get the result. From Lemma~\ref{lem:l1_l2_bound_pruning} (with $s_1 = s, \mathcal{S} = \hat{\T}_{l,k}, s_2 = s, \|\mbe\| \leq \epsilon$), we have
\begin{eqnarray*}
\|\mbx-\hat{\mbx}_{l,k}\| \leq \sqrt{\frac{1}{1-\delta_{2s}^2}}\|\mbx_{\hat{\T}^c_{l,k}}\| + \frac{\sqrt{1+\delta_{s}}}{1-\delta_{2s}}\epsilon.
\end{eqnarray*}
Also from Lemma~\ref{lem:smaller_indices_bound_l1_l2} (with $s_1 = s, \mbz = \tilde{\mbx}_{l,k}, s_2 = N, \mathcal{S}_{\nabla} = \hat{\T}_{l,k}^c$), we can write
\begin{eqnarray*}
\|\mbx_{\hat{\T}^c_{l,k}}\| \leq \sqrt{2}\|\mbx - \tilde{\mbx}_{l,k}\|.
\end{eqnarray*}
We get the $\ell_2$-bound by combining the above two equations. Next consider the following $\ell_1$ bound derived in Lemma~\ref{lem:l1_l2_bound_pruning} (with $s_1 = s, \mathcal{S} = \hat{\T}_{l,k}, s_2 = s, \|\mbe\| \leq \epsilon$)
\begin{eqnarray*}
\|\mbx-\hat{\mbx}_{l,k}\|_1 \leq \sqrt{\frac{2s}{1-\delta_{2s}^2}}\|\mbx_{\hat{\T}^c_{l,k}}\|_1 + \frac{\sqrt{2ss(1+\delta_{s})}}{1-\delta_{2s}}\epsilon.
\end{eqnarray*}
Also, from Lemma~\ref{lem:smaller_indices_bound_l1_l2} (with $s_1 = s, \mbz = \tilde{\mbx}_{l,k}, s_2 = N, \mathcal{S}_{\nabla} = \hat{\T}_{l,k}^c$), we have
\begin{eqnarray*}
\|\mbx_{\hat{\T}^c_{l,k}}\|_1 \leq \|\mbx - \tilde{\mbx}_{l,k}\|_1.
\end{eqnarray*}
Now the $\ell_1$-bound follows from combining the above two equations.
\QEDB

\subsection{Proof of Theorem~\ref{thm:l1_pruning_theorem_convergence}}
Under the assumption $\|\mbx\|_0 = s$, we have from Theorem~\ref{thm:l1_conv_theorem1},
\begin{eqnarray*}
\|\mbx-\tilde{\mbx}_{l,k}\| \leq \frac{4\lambda \sqrt{s(1+\delta_{s+a})}}{c_1(2\lambda -1)}\epsilon .
\end{eqnarray*}
Plugging the above inequality in Lemma~\ref{lem:pruning_bound_lemma1}, we get the first bound.

For the second part, we have to derive the recurrence inequality. From Theorem~\ref{thm:l1_recurrence_theorem2}, we have
\begin{eqnarray*}
\|\mbx-\tilde{\mbx}_{l,k}\|_1 \leq c_2 \sum\limits_{r \in \mathcal{N}_l} h_{lr} \|\mbz_{r,k-1}\|_1 + c_4 \epsilon.
\end{eqnarray*}
Using the second bound from Lemma~\ref{lem:pruning_bound_lemma1} in the above equation, we get
\begin{eqnarray*}
\begin{array}{l}
\|\mbz_{l,k}\|_1  \leq \|\mbx-\hat{\mbx}_{l,k}\|_1 \leq \sqrt{\frac{2s}{1-\delta_{2s}^2}}\|\mbx-\tilde{\mbx}_{l,k}\|_1 + \frac{\sqrt{2s(1+\delta_{s})}}{1-\delta_{2s}}\epsilon \\
 \hspace{3mm} \leq \frac{c_{2}\sqrt{2s}}{\sqrt{1-\delta_{2s}^2}}\sum\limits_{r \in \mathcal{N}_l} h_{lr} \|\mbz_{r,k-1}\|_1 + \left(\frac{c_{4}\sqrt{2s}}{\sqrt{1-\delta_{2s}^2}} + \frac{\sqrt{2s(1+\delta_{s})}}{1-\delta_{2s}}\right)\epsilon.
\end{array}
\end{eqnarray*}
The initialization step error at node $l$ can be bounded as $\|\mbx-\hat{\mbx}_{l,0}\|_1 \leq \left(c_{7} \sqrt{\frac{2s}{1-\delta_{2s}^2}} + \frac{\sqrt{2s(1+\delta_{s})}}{1-\delta_{2s}}\right) \epsilon$ by applying $\lambda = 1$ to the above equation.
We can now follow the proof of Theorem~\ref{thm:l1_convergence_theorem2} to show that the error at iteration $k$ is bounded as
\begin{eqnarray*}
\begin{array}{l}
\hspace{-1mm}\underline{\|\mbz_k\|_1} \\ \leq \frac{c_{14}^k-1}{c_{14}-1}\left((c_{4}+c_{14}c_{7}) \sqrt{\frac{2s}{1-\delta_{2s}^2}} + \frac{(1+c_{14})\sqrt{2s(1+\delta_{s})}}{1-\delta_{2s}}\right) \epsilon \mathbf{1}_L,
\end{array}
\end{eqnarray*}
where $c_{14} = c_2\sqrt{\frac{2s}{1-\delta_{2s}^2}}$. 
The error bound at the individual nodes follows from the above inequality.  \QEDB

\subsection{Proof of Theorem~\ref{thm:l2_pruning_theorem_convergence}}
Under the assumption $\|\mbx\|_0 = s$, we have from Theorem~\ref{thm:l2_conv_theorem1},
\begin{eqnarray*}
\|\mbx-\tilde{\mbx}_{l,k}\| \leq \frac{4\lambda \sqrt{s(1+\delta_{s+a})}}{c_1(2\lambda -1)} \epsilon .
\end{eqnarray*}
Plugging the above inequality in Lemma~\ref{lem:pruning_bound_lemma1}, we get the first bound.

For the second part, we have to derive the recurrence inequality. From Theorem~\ref{thm:l2_recurrence_theorem2}, we have
\begin{eqnarray*}
\|\mbx-\tilde{\mbx}_{l,k}\| \leq c_9 \sum\limits_{r \in \mathcal{N}_l} h_{lr} \|\mbz_{r,k-1}\| + c_{11} \epsilon.
\end{eqnarray*}
Using the first bound from Lemma~\ref{lem:pruning_bound_lemma1} in the above equation, we get
\begin{eqnarray*}
\begin{array}{l}
\|\mbz_{l,k}\|  \leq \|\mbx-\hat{\mbx}_{l,k}\| \leq \sqrt{\frac{2}{1-\delta_{2s}^2}}\|\mbx-\tilde{\mbx}_{l,k}\| + \frac{\sqrt{1+\delta_{s}}}{1-\delta_{2s}}\epsilon \\
 \hspace{3mm} \leq c_9\sqrt{\frac{2}{1-\delta_{2s}^2}}\sum\limits_{r \in \mathcal{N}_l} h_{lr} \|\mbz_{r,k-1}\| + \left(c_{11} \sqrt{\frac{2}{1-\delta_{2s}^2}} + \frac{\sqrt{1+\delta_{s}}}{1-\delta_{2s}}\right)\epsilon.
\end{array}
\end{eqnarray*}
The initialization step error at node $l$ can be bounded as $\|\mbx-\hat{\mbx}_{l,0}\| \leq \left(c_{13} \sqrt{\frac{2}{1-\delta_{2s}^2}} + \frac{\sqrt{1+\delta_{s}}}{1-\delta_{2s}}\right) \epsilon$ by applying Lemma~\ref{lem:pruning_bound_lemma1} to Lemma~\ref{lem:BPDN_convergence_result}.
We can now follow the proof of Theorem~\ref{thm:l2_convergence_theorem2} to show that the error at iteration $k$ is bounded as
\begin{eqnarray*}
\begin{array}{l}
\hspace{-1mm}\underline{\|\mbz_k\|} \leq \frac{c_{15}^k-1}{c_{15}-1}\left((c_{11}+c_{15}c_{13}) \sqrt{\frac{2}{1-\delta_{2s}^2}} + \frac{(1+c_{15})\sqrt{1+\delta_{s}}}{1-\delta_{2s}}\right) \epsilon \mathbf{1}_L,
\end{array}
\end{eqnarray*}
where $c_{15} = c_9\sqrt{\frac{2}{1-\delta_{2s}^2}}$. The error bound at the individual nodes follows from the above inequality.  \QEDB





%
\bibliographystyle{ieeetr}
\bibliography{references/ref_zaki,references/biblio_saikat_CS1,references/biblio_saikat_Pub,references/biblio_saikat_BigData,references/ref_distributed_sparse}


\end{document}